\documentclass{article}

\PassOptionsToPackage{sort}{natbib}

 \usepackage[main, final]{neurips_2026}
 
\usepackage[utf8]{inputenc} 
\usepackage[T1]{fontenc}    
\usepackage{hyperref}       
\usepackage{url}            
\usepackage{booktabs}       
\usepackage{amsfonts}       
\usepackage{nicefrac}       
\usepackage{microtype}      

\usepackage{wrapfig}        
\usepackage{graphicx}       
\usepackage{amsmath}
\usepackage{multirow}
\usepackage[table]{xcolor}
\usepackage{stfloats}
\usepackage{float}
\usepackage{enumitem}
\usepackage{makecell}
\usepackage{lipsum, comment}
\usepackage{caption}
\usepackage{subcaption}

\definecolor{best}{rgb}{0.8, 0.9, 1.0}
\definecolor{2ndbest}{rgb}{0.9, 1.0, 0.8}

\makeatletter
\usepackage{xspace}
\def\@onedot{\ifx\@let@token.\else.\null\fi\xspace}
\DeclareRobustCommand\onedot{\futurelet\@let@token\@onedot}

\newcommand{\eqnref}[1]{Eq\onedot~\eqref{#1}}
\newcommand{\figref}[1]{Fig\onedot~\ref{#1}}

\newcommand{\secref}[1]{Section~\ref{#1}}
\newcommand{\tabref}[1]{Tab\onedot~\ref{#1}}
\newcommand{\appref}[1]{Appendix~\ref{#1}}

\def\eg{\emph{e.g}\onedot}
\def\ie{\emph{i.e}\onedot}

\usepackage{pifont}

\newcommand{\cmarkB}{\textcolor{blue!30!cyan}{\ding{51}}} 
\newcommand{\xmark}{\textcolor{red}{\ding{55}}}

\newcommand{\mycol}{\cellcolor{violet!70!magenta!10}}

\newcommand{\colcyanA}{\cellcolor{blue!20!cyan!10}}
\newcommand{\mycolB}{\cellcolor{gray!10}}

\newcommand{\mycolC}{\cellcolor{lime!35}}
\newcommand{\mycolD}{\cellcolor{green!25}}

\title{UDT: Reconciling U-Nets and Diffusion Transformers with Data-Adaptive Token Reduction}

\author{
  Junno Yun \\
  University of Minnesota\\
  \texttt{yun00049@umn.edu} \\
  \And
  Ya\c{s}ar Utku Al\c{c}alar \\
  University of Minnesota\\
  \texttt{alcal029@umn.edu} \\
  \And
  Mehmet Ak\c{c}akaya\thanks{Corresponding Author} \\
  University of Minnesota\\
  \texttt{akcakaya@umn.edu}
}

\begin{document}

\maketitle

\begin{abstract}
Diffusion Transformers (DiTs) have emerged as a core architecture in generative modeling due to their scalability and adaptability to multimodal tasks. DiTs comprise isotropic transformer blocks, and learn representations progressively across depth, where the denoising objective drives later layers to focus on fine-detail reconstruction. This results in degraded representation quality and an imbalanced encoder--decoder behavior. Prior approaches such as representation alignment (REPA) mitigate this by encouraging stronger early representations via training regularization. Alternatively, U-Net-style DiT architectures introduce explicit multi-scale encoder–decoder structures for improved convergence. But they build on standard U-Net wisdom via learnable operators for spatial downsampling, which are not well-suited to transformer architectures, introducing inefficiencies and compatibility issues with components such as cross-attention and representation regularization. In this work, we propose \textbf{UDT}, a U-Net diffusion transformer that combines the representation power of DiTs with the encoding–decoding benefits of U-Nets, through data-adaptive token merging for downsampling and upsampling, while preserving the DiT token dimension. Our baseline UDT architecture outperforms existing U-Net DiTs and achieves performance comparable to REPA across all model sizes. Furthermore, using architectural optimization and REPA, UDT outperforms SiT's 7.9 FID at 1400 epochs (w/o CFG) within 40 epochs ($\sim 40\times$ faster convergence) for XL model size on $256 \times 256$ ImageNet. Finally, it achieves strong image generation performance with CFG, reaching FID of 1.38 (320 epochs) with SD-VAE and 1.35 (500 epochs) with VA-VAE, providing a new backbone for DiTs with strong empirical benefits. Code is available at \url{https://github.com/JN-Yun/UDT}.  
\end{abstract}

\section{Introduction} \label{sec:introduction}
Vision Transformers (ViTs)~\cite{dosovitskiy2021an} extend the transformer architecture~\cite{vaswani2017attention} to image data by representing 2D images as sequences of patches and applying self-attention to capture global contextual relationships. With the rapid advancement of generative modeling, transformer-based architectures have been increasingly adopted in diffusion models (DMs), leading to the development of diffusion transformers~\cite{bao2023uvit, peebles2023dit, ma2024sit, tian2024udit, Zheng2024MaskDiT}. Among these, DiT~\cite{peebles2023dit} has played a pivotal role by introducing a highly scalable architecture that effectively leverages transformer design for diffusion modeling, while SiT~\cite{ma2024sit} extends this paradigm to the rectified flow framework~\cite{lipman2023flow}. Their flexibility and strong performance have established diffusion transformers as powerful backbones not only for image generation but also for broader applications such as text-to-image (T2I) generation~\cite{esser2024mmdit, chen2024pixart, xie2024sana, wang2025lit}.

From a representational perspective, it is well understood that diffusion transformers inherently learn semantically meaningful representations through the denoising objective~\cite{baranchuk2022labelefficient, xiang2023DDAE, chen2025deconstructing}. Since diffusion transformers are based on the ViT architecture, which consists of a sequence of isotropic transformer blocks, they naturally learn representations progressively across depth, with semantic representation quality gradually improving from early to later layers~\cite{li2023mage}. However, diffusion models must ultimately focus on denoising and fine-detailed reconstruction, which requires recovery of higher-frequency information in later stages. As a result, representation quality tends to peak in the middle-to-late layers and then decrease toward the final layers, where the model shifts its focus from semantic encoding to denoising generation~\cite{xiang2023DDAE, yu2025REPA}. This creates an imbalanced encoder--decoder structure, where the encoding stage becomes relatively deep while the effective decoding stage remains short. To address this, a line of work aims to encourage earlier formation of strong representations through training regularization, effectively moving the encoder forward and allowing later layers to devote more capacity to denoising. This direction has been explored through representation alignment with external encoders~\cite{yu2025REPA, leng2025REPAE}, promoting linear separability~\cite{yun2025_LSEP}, self-contrastive learning methods~\cite{wang2025diffuse}, and improving latent representations through stronger variational autoencoder (VAE) designs~\cite{leng2025REPAE,yao2025reconstruction, zheng2026RAE}.

\begin{figure*}[t]
    \centering
    \includegraphics[width=0.98\linewidth]{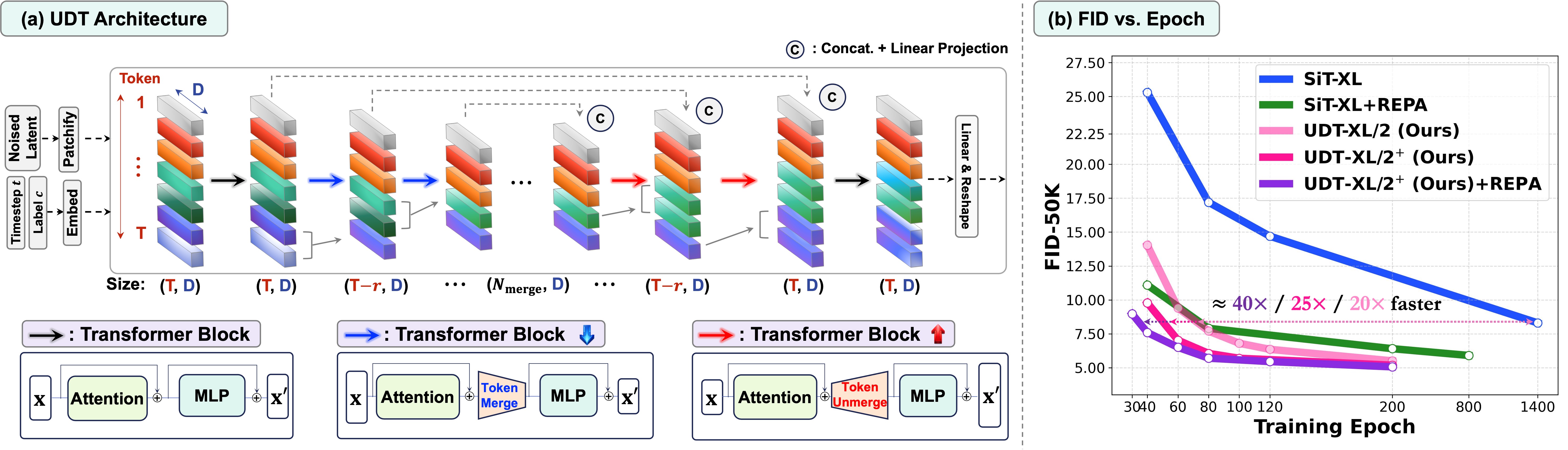} 
    \vspace{-0.1cm}
    \caption{(a) Architecture of the proposed method \textbf{UDT}. It preserves the token hidden dimension ($D$) while gradually reducing and restoring the token sequence length via data-adaptive token merging and unmerging. (b) FID vs. Epoch on XL models for ImageNet $256 \times 256$ without classifier-free guidance. UDT (our baseline, marked in light pink) and UDT$^+$ (baseline + advanced techniques, marked in pink), achieve 7.7 FID at 80 epochs and 7.0 FID at 60 epochs, \emph{without additional regularization or VAE modification}, outperforming SiT-XL's 7.9 FID at 1400 epochs. Our method with REPA achieves 7.6 FID at \emph{just} 40 epochs, converging significantly faster than the SiT baseline and REPA.}
    \vspace{-0.8cm}
    \label{fig:overview}
\end{figure*}

Explicit encoder--decoder architectures naturally arise in CNN-based U-Net~\cite{ronneberger2015u} backbones that integrate spatial downsampling and upsampling with increasing channel dimensions and skip connections, improving gradient flow and preserving low- and high-frequency information. DDPM~\cite{ho2020ddpm} adopts a CNN-based U-Net for diffusion modeling, and subsequent works~\cite{dhariwal2021beatGANs, karras2022elucidating} have established U-Net variants as the standard backbone, motivating their adaptation to diffusion transformers~\cite{bao2023uvit, tian2024udit, tian2026urepa}. U-ViT~\cite{bao2023uvit} incorporated skip connections into a ViT-based diffusion transformer without explicit spatial resolution changes, while later works, including U-DiT~\cite{tian2024udit} and SiT$\downarrow$ in UREPA~\cite{tian2026urepa}, showed that the fast-convergence benefit primarily arises from multi-scale hierarchical modeling enabled by downsampling, with skip connections remaining important for mitigating information loss from reduced spatial resolution. However, their downsampling process relies on fixed local neighborhoods, ignoring similarity between tokens across the image, thereby weakening token-wise interactions in self-attention. It also complicates patch-wise representation alignment~\cite{yu2025REPA}, requiring additional adaptation modules and regularization designs~\cite{tian2026urepa}. Moreover, architectures such as U-DiT vary channel dimensions across layers, introducing representation inconsistencies and additional engineering overhead for extensions such as T2I generation with cross-attention.

\noindent \textbf{Our approach:} 
Building on these observations, we are led to a fundamental question:

{\centering
\emph{“Can we introduce an encoder-decoder architecture into diffusion transformers without disrupting their representational characteristics and self-attention dynamics?”}\par}

Our design is guided by two key principles: (1) While existing U-Net DiTs reduce the channel dimension of tokens to achieve spatial downsampling, we instead perform no dimensionality reduction in this direction and preserve the channel dimension (\ie, token feature dimension) to maintain representation quality. This design also improves compatibility with downstream tasks such as T2I generation and representation alignment. (2) Rather than relying on fixed, learnable operators for spatial reduction, we advocate for a data-adaptive mechanism that accounts for token importance, enabling selective compression of similar or less informative regions while retaining critical details.

To this end, we propose \textbf{UDT}, a U-Net diffusion transformer, which introduces a transformer-specific encoder--decoder architecture that incorporates token merging (ToMe)~\cite{bolya2022tome} for data-adaptive and training-free downsampling and upsampling. Following the U-Net paradigm of skip connections and downsampling/upsampling (\figref{fig:overview}(a)), semantically similar tokens (\eg, background regions) are progressively merged based on feature similarity, with merge indices recorded to enable exact unmerging in the decoder, while skip connections mitigate information loss induced by spatial reduction. This design adapts token merging as a transformer-compatible mechanism for resolution reduction, enabling an effective encoder--decoder structure, facilitating hierarchical representation learning while improving computational efficiency. Experiments show that UDT, out-of-the-box, substantially improves both training efficiency and generative performance on flow-based transformer models such as SiT~\cite{ma2024sit}, \emph{without requiring any additional regularization or modifications to model configurations}. Our main contributions are as follows:
\begin{itemize}[leftmargin=*, itemsep=0em, topsep=0pt]
    \item We introduce a novel architecture, \textbf{U-Net Diffusion Transformer (UDT)}, which provides a new perspective on downsampling and upsampling in U-shape transformers via a data-adaptive token merging mechanism while preserving the token representation dimension.
    \item UDT reduces computational cost by progressively decreasing the token sequence length in an adaptive manner. Despite its lightweight architecture, our XL model achieves 7.7 FID out-of-the-box at only 80 epochs without CFG, improving on the baseline SiT model's 7.9 FID at 1400 epochs ($\sim 20\times$ faster convergence), while also outperforming even REPA methods. UDT variants with advanced techniques and REPA further improve generative performance, achieving 7.6 FID at 40 epochs ($\sim 40\times$ faster). Using CFG on 256$\times$256 ImageNet, UDT achieves FID scores of 1.38 (at 320 epochs) with SD-VAE and 1.35 (at 500 epochs) with VA-VAE, where samples are generated by randomly sampling class labels, \ie without class-balanced sampling.
    \item We adapt UDT for longer token sequences by leveraging early-stage token merging. By using high token reduction rates, suitable for such long sequences of patches, in the first few encoder blocks, followed by gradual merge/unmerge in later stages, efficient computation is achieved for high-resolution or long-sequence settings (\eg, patch size 1 or higher resolution at $512 \times 512$) without architectural modification.
    \item Our improvements stem purely from architectural design, while retaining the same configuration as the DiT baseline. Thus, UDT can serve as a drop-in replacement for various DiT variants, including DiT~\cite{peebles2023dit} with $\epsilon$ prediction, pixel-space DiT such as JiT~\cite{li2025jit}, models using improved VAE architectures such as VA-VAE~\cite{yao2025reconstruction}, and T2I models such as MMDiT~\cite{esser2024mmdit}, demonstrating faster convergence and strong adaptability.
\end{itemize}


\section{Related Work} \label{sec:related_work}

\subsection{Representation Learning in Diffusion Transformers} \label{sec:representation_learning}
To address the characteristic of diffusion transformers where the encoding stage becomes relatively deep while the effective decoding stage remains short, a line of work has aimed to encourage the earlier formation of strong representations, effectively moving the encoder forward and allowing later layers to devote more capacity to denoising.

One prominent direction is representation alignment ~\cite{yu2025REPA, tian2026urepa,leng2025REPAE,yao2025reconstruction,jiang2025SREPA}. Among these, REPA~\cite{yu2025REPA, tian2026urepa} improves training efficiency and generative performance by leveraging high-quality features from pretrained self-supervised visual encoders, such as DINOv2~\cite{oquab2024dinov}, and aligning them with early intermediate representations of diffusion transformers. Subsequent variants extend alignment with external encoders into the VAE latent space~\cite{leng2025REPAE,yao2025reconstruction,zheng2026RAE}. Other approaches, such as Self-REPA (SRA)~\cite{jiang2025SREPA}, avoid reliance on large-scale external encoders via internal alignment by using deeper layers, which naturally exhibit stronger representations, as teachers for earlier layers. In addition, self-supervised objectives such as self-contrastive learning~\cite{wang2025diffuse} and methods that promote linear separability~\cite{yun2025_LSEP} improve early-layer representations without external supervision.

These studies consistently show that strengthening representations in the early stages of the model, thereby allowing subsequent layers to focus more on denoising, can significantly accelerate convergence and improve generative quality. However, these approaches \emph{do not fundamentally improve representation learning through the diffusion transformer architecture itself}. Instead, they rely on additional external visual encoders or modifications to the VAE, which are not always applicable, or on auxiliary regularization terms, which naturally introduce computational overhead during training.

\begin{figure*}[t]
    \centering
    \includegraphics[width=1.0\linewidth]{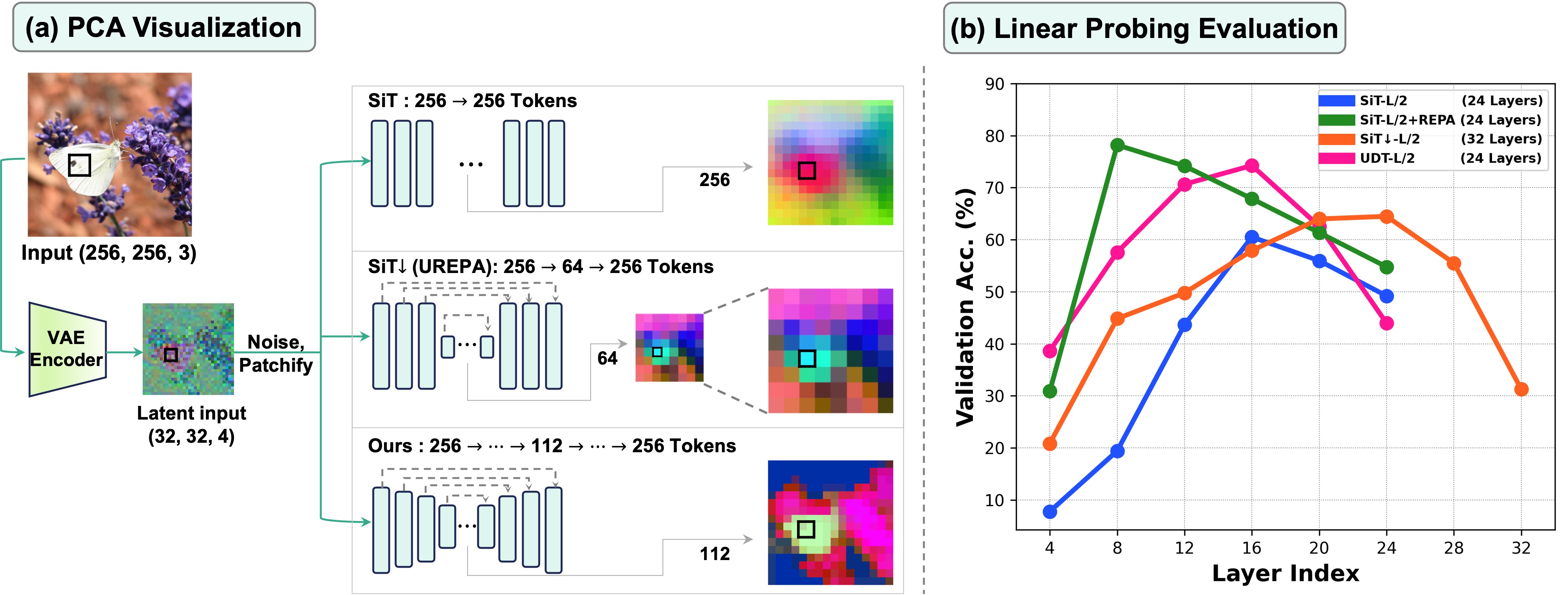} 
    \vspace{-0.2cm}
    \caption{\textbf{Representation Analysis.} (a) PCA visualization of intermediate layers. Features are extracted from the bottleneck layer (\eg, layer 18 of 24 for SiT-L/2, layer 16 of 32 for SiT$\downarrow$, and layer 12 of 24 for UDT). Features with 112 tokens in our method are unmerged for visualization. (b) Linear probing evaluation on pretrained models across layers. All experiments are conducted at noise level $t=0.1$ using Large models trained for 80 epochs on ImageNet $256 \times 256$.}
    \label{fig:representation_vis}
    \vspace{-.5cm}
\end{figure*}

\subsection{U-Shape Diffusion Transformers} \label{sec:unet_dit}
Motivated by the structural advantages of the U-Net~\cite{ronneberger2015u} architecture, encoder--decoder-style designs have been explored for DiTs. U-ViT used skip connections without down/upsampling, while later works such as U-DiT~\cite{tian2024udit} and SiT$\downarrow$ in UREPA~\cite{tian2026urepa} introduce explicit resolution changes, enabling hierarchical modeling. Empirically, these works attribute improved optimization and convergence to these multi-scale representations, while skip connections facilitate preserving information lost during spatial compression. Both designs use fixed $2\times2$ spatial grid downsampling, where the dimensionality of tokens are first reduced using a learnable convolutional kernel. Subsequently, four spatially neighboring tokens are grouped into a single token by stacking them along the channel dimension, thereby reducing the token sequence length, \ie spatial dimension. This use of fixed local neighborhoods weakens token-wise interactions in self-attention. Implementation-wise, U-DiT scales model capacity with increased feature dimensions, while SiT$\downarrow$ increases number of transformer blocks. Both models further use architectural refinements, not present in original DiT/SiT. U-DiT adopts cosine similarity attention~\cite{liu2021swin}, rotary positional embeddings (RoPE)~\cite{su2024roformer}, depth-wise convolutional FFN~\cite{wang2022uformer}, and reparameterized FFN~\cite{ding2021repvgg}, while SiT$\downarrow$ employs SwiGLU~\cite{shazeer2020glu} and RoPE~\cite{su2024roformer}.

Despite their benefits, existing U-Net DiTs use fixed-neighborhood down/upsampling schemes~\cite{tian2024udit, tian2026urepa}, which introduce a fundamental tension within the transformer paradigm. These methods first compress token features and then spatially concatenate neighboring tokens along the channel dimension. While this preserves local information, it reduces four tokens into one \emph{without} considering their global semantic similarity, thereby diminishing token-level diversity and limiting token-wise interaction capacity in self-attention to model global dependencies (\figref{fig:representation_vis}(a), middle). For instance, a single downsampling in SiT$\downarrow$ reduces 256 tokens to 64, significantly constraining self-attention. This highlights the need for downsampling that balances computational efficiency with preserving token importance and attention-critical spatial information. Finally, existing spatial downsampling schemes complicate representation alignment with pretrained ViT encoders~\cite{tian2026urepa}, due to token dimensionality mismatch (\eg, 64 vs.\ 256 tokens), hindering direct token-wise alignment~\cite{tian2026urepa}.

\subsection{Token Merge} \label{sec:tome}
Token pruning~\cite{rao2021dynamicvit, meng2022adavit, kong2022spvit, yin2022vit, liang2022not} and token merging~\cite{bolya2022tome, marin2021token, ryoo2021tokenlearner} have recently emerged as a promising subfield of ViTs. These methods harness the input-agnostic nature of transformers and the widely observed redundancy of tokens in ViTs for downstream prediction. Based on this property, redundant tokens can be removed or merged at runtime, enabling faster inference with minimal performance degradation while achieving significant computational savings.

Among them, Token Merging (ToMe)~\cite{bolya2022tome} progressively merges tokens, reducing the number of tokens by $r$, thereby accelerating subsequent blocks through several key components: (1) Tokens are merged across layers based on the similarity between the keys of each token. (2) Efficient token reduction is achieved via bipartite soft matching, which partitions tokens into two disjoint sets and connects each token in one set to its most similar counterpart in the other, forming a sparse bipartite graph. Only the top-$r$ most similar connections are retained, and connected token pairs are merged via weighted averaging. (3) Two similar tokens with channel dimension $D$, $\mathbf{x}_1, \mathbf{x}_2 \in \mathbb{R}^D$,  are merged into a single token using a weighted average based on token size:
\begin{equation}
\mathbf{x}_{1,2}^{\textrm{m}} = \big(\textrm{s}_1 \mathbf{x}_1 + \textrm{s}_2 \mathbf{x}_2\big)/\big(\textrm{s}_1 + \textrm{s}_2\big),
\label{eq:tome_merge}
\end{equation}
where $\textrm{s}_1$ and $\textrm{s}_2$ denote the number of original patches represented by each token. The merged token $\mathbf{x}_{1,2}^{\textrm{m}}$ is associated with the combined size $\textrm{s} = \textrm{s}_1 + \textrm{s}_2$. 
(4) Lastly, since merged tokens no longer correspond to a single input patch, proportional attention is applied to correct the bias introduced by merging as: $\mathrm{softmax}(\mathbf{QK}^T/\sqrt{d} + \log \mathbf{s}),$ where $\mathbf{s}$ is a row vector representing the number of original patches each token corresponds to. This formulation is equivalent to replicating each merged token $s$ times in the attention computation. 

ToMe also allows approximate recovery of the original token layout when merge indices are recorded. This unmerging step is crucial for dense prediction tasks as in DMs, which require predicting noise for every original token~\cite{bolya2023token}. During unmerging, the merged token is copied back to the original token positions, while the associated sizes are restored to $\textrm{s}_1$ and $\textrm{s}_2$, respectively. Across various pretrained ViT models such as MAE~\cite{he2022masked}, even with over 95\% token reduction, token merging can maintain classification accuracy with only a negligible drop while enabling faster inference, while preserving meaningful semantic information~\cite{bolya2022tome}. We note that beyond ViTs for classification, ToMe has been applied to pretrained DMs~\cite{bolya2023token} by merging tokens before self-attention for faster inference and reduced memory. But its utility has not been explored for improving training dynamics.

\section{UDT: U-Net-based Diffusion Transformer with Token Merge}
Building on these prior ideas, we propose a U-Net-type diffusion transformer with token merging. The proposed architecture adopts a U-Net structure and introduces transformer-friendly downsampling and upsampling via data-adaptive, parameter-free token merging and unmerging, while preserving the original DiT token representation dimension. Thus, dimensionality reduction is only constrained to the spatial dimension across tokens, without modifying token lengths themselves.

\begin{figure*}[t]
    \centering
    \includegraphics[width=1.0\linewidth]{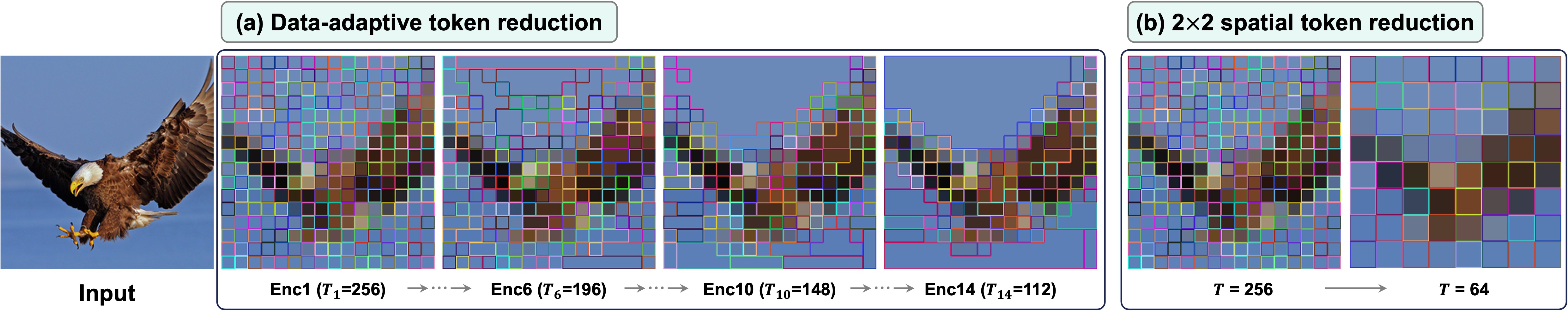} 
    \vspace{-0.3cm}
    \caption{Visualization of token reduction: (a) Data-adaptive token reduction (ours) gradually merges semantically similar tokens (e.g., background regions), allowing fine-grained details to be retained. (b) Spatial token reduction with factor-2 downsampling, which relies on fixed local neighborhoods and ignores similarity between tokens across the image, resulting in the loss of fine details.}
    \label{fig:data_adaptive_vs_spatial}
    \vspace{-3ex}
\end{figure*}

\textbf{U-Net shape architecture.} We first design a U-Net-shaped diffusion transformer, as illustrated in~\figref{fig:overview}(a), while preserving the original DiT configurations, including the same number of blocks, attention heads, and hidden dimensions, as detailed in~\appref{appdx:implmentation_detail}. The entire network is symmetrically divided into encoder and decoder stages:
\begin{itemize}[leftmargin=*, itemsep=0em, topsep=0pt]
    \item The first encoder block and the final decoder block maintain the full token resolution and are connected through a skip connection, since diffusion models require noise prediction for every original token. This ensures that the input and output stages operate on full-resolution tokens without being affected by downsampling or upsampling (\ie token merging and unmerging).
    \item From the second encoder block to the block before the bottleneck, token merging is progressively applied, reducing the number of tokens by $r$ at each block until reaching the target number of merged tokens, denoted by $N_{\textrm{Merge}}$ (see~\appref{appdx:implmentation_detail} for details). 
    \item The bottleneck consists of two middle blocks operating at the lowest token resolution. 
    \item In the decoder, token unmerging is progressively performed by restoring $r$ tokens at each block using the recorded merge indices from the corresponding encoder stages.    
\end{itemize}
As shown in~\tabref{tab:ablation_tome}(a), performance peaks at $N_{\textrm{Merge}}=112$ for our base model, with consistent behavior observed across other model sizes. An overview of the architecture, including token merging in the encoder and symmetric unmerging in the decoder, is shown in Fig.~\ref{fig:overview}(a).

\textbf{Down/Upsampling with data-adaptive token merge/unmerge.}
We examine the effectiveness of data-adaptive token merging for down/upsampling by comparing it to learnable projection operators, replacing token merge/unmerge in Fig.~\ref{fig:overview}(a) with learnable projections within the same architecture. In this variant, the down/upsampling modules in~\figref{fig:overview}(a) are replaced with a learnable projection function instead of token merge/unmerge. Given an input feature $\mathbf{x}^{(i)} \in \mathbb{R}^{T_i \times D}$, we apply a learnable projection $g_i(\cdot)$ that reduces the token length by a fixed amount $r$, \ie, $ g_i: \mathbb{R}^{T_i \times D} \rightarrow \mathbb{R}^{(T_i - r) \times D}. $ This operation is applied progressively across layers and symmetrically used for upsampling. While this design allows gradual token reduction, it introduces token mixing, which leads to a significant degradation in performance as shown in \tabref{tab:ablation_tome}(b). Similarly, comparing representations of our strategy to the downsampling scheme that includes compression in token dimensions used in prior U-shaped diffusion transformers (\eg, U-DiT and SiT$\downarrow$) highlights that the latter may blur fine spatial details by aggressively compressing local structures as illustrated in~\figref{fig:representation_vis}(a) (middle), showing PCA visualizations of bottleneck features. Our method, instead, enables flexible token reduction through data-adaptive token merging, where semantically similar tokens (\eg, background regions) are preferentially merged~(\figref{fig:data_adaptive_vs_spatial}). As shown in~\figref{fig:representation_vis}(a) (bottom), this allows fine-grained details (\eg, textures and small structures) to be retained, while redundant regions are effectively compressed. Additionally, since the token count is gradually reduced throughout the network rather than being abruptly compressed at a single stage, inference efficiency is also improved across the entire model.

\textbf{Adapting general transformer improvements.}
A wide range of techniques have been proposed to improve transformers~\cite{liu2021swin, su2024roformer, ding2021repvgg, wang2022uformer, shazeer2020glu, esser2024mmdit}, and many of these have also shown effectiveness in diffusion transformers~\cite{tian2024udit, esser2024mmdit,tian2026urepa,li2025jit}. Building on this, our method incorporates such improvements to enhance both training efficiency and generative performance. In particular, we adopt techniques with minimal impact on parameters and FLOPs, including log-normal time sampling~\cite{esser2024mmdit}, RoPE~\cite{su2024roformer}, and SwiGLU~\cite{shazeer2020glu}. For RoPE, it is applied only to the first and last blocks, since token merging/unmerging disrupts the original positional indices and restricts its use to stages where the token count remains 256. We denote UDT as the baseline model and UDT$^+$ as the version incorporating these techniques.

\begin{table*}[!t]
\centering
\small
\caption{\textbf{Different Token Merging Setups.} (a) Optimal number of tokens at the bottleneck. (b) Comparison between gradual downsampling with learnable operators and token merging. (c) Effects of different components of token merging. All experiments are performed with the B/2 model on ImageNet $256 \times 256$ without CFG (80 epochs).}
\vspace{-2ex}
\label{tab:ablation_tome}
\vspace{+0.2cm}
\setlength{\tabcolsep}{2pt}

\resizebox{\textwidth}{!}{%
\begin{tabular}{ccc}
\renewcommand{\arraystretch}{0.7}
\begin{tabular}{cc}
\toprule
\mycol $N_\textbf{Merge}$ & \mycol FID$\downarrow$ \\
\midrule
96 & 24.3 \\
\textbf{112} & \textbf{24.0} \\
128 & 24.2 \\
144 & 24.3 \\
160 & 25.1 \\
\bottomrule
\end{tabular}

&

\renewcommand{\arraystretch}{1.1}
\begin{tabular}{ccc}
\toprule
\mycol Downsample Method & \mycol \# Tokens & \mycol FID$\downarrow$ \\
\midrule

SiT-B/2 & 256 & 33.0 \\
\arrayrulecolor{gray!50}
\cmidrule(lr){1-3}

Learnable Oper. & 256 $\rightarrow$ 112 & 45.5 \\
Data-Adaptive & 256 $\rightarrow$ 112 & \textbf{24.0} \\

\arrayrulecolor{black}
\bottomrule
\end{tabular}

&

\renewcommand{\arraystretch}{1.}
\begin{tabular}{cccc}
\toprule
\mycol Key-based Sim. &
\mycol Weighted Avg. &
\mycol Proport. Attn. &
\mycol FID$\downarrow$ \\
\midrule

\xmark & \xmark & \xmark & 25.9 \\
\cmarkB &        &        & 24.5 \\
\cmarkB & \cmarkB &       & 24.2 \\
\cmarkB & \cmarkB & \cmarkB & \textbf{24.0} \\

\bottomrule
\end{tabular}

\\

\footnotesize (a) 
&
\footnotesize (b) 
&
\footnotesize (c) 

\end{tabular}%
}
\vspace{-0.5cm}

\end{table*}

\textbf{Extending to REPA~\cite{yu2025REPA}.}
Given its effectiveness in improving training efficiency and generative performance, compatibility with REPA is a desirable property for diffusion transformer architectures. As expected from CNN-based U-Net intuition, UREPA~\cite{tian2026urepa} shows that the bottleneck of a U-Net diffusion transformer is the most suitable location for representation alignment. However, fixed spatial downsampling~\cite{tian2024udit, tian2026urepa} introduces a mismatch with external ViT encoders, limiting faithful patch-wise alignment and requiring additional components such as upsampling MLPs and manifold losses. In contrast, our method does not require such additional engineering designs and remains compatible with the standard REPA setting. Since external encoder features share the same token resolution as the model input, we restore merged bottleneck features to the original resolution using the recorded merge indices and the unmerging operation. This allows us to recover features at the input token size from the bottleneck, where alignment is performed. Representation alignment is then achieved using an MLP projection, enabling direct patch-wise alignment without additional spatial adaptation modules or regularization loss (\eg, manifold loss)~\cite{tian2026urepa}. We selected the last encoder layer as the target layer for REPA.

\section{Experiments} \label{sec:experiment}

\clearpage

\subsection{Setup}
\textbf{Implementation details.}

\begin{table}[t]
\centering

\begin{minipage}{0.35\textwidth}
\vspace{-0.2cm}
\caption{FID comparison, generated without CFG. ${\dagger}, +$: using selected advanced techniques.}
\vspace{-0.1cm}
\label{tab:FID_comparison}
\vspace{0.1cm}
\centering

\begin{subtable}{\linewidth}
\centering
\centering
\scriptsize
\setlength{\tabcolsep}{4.0pt}
\renewcommand{\arraystretch}{1.37}

\vspace{+0.1cm}
\begin{tabular}{@{}ccc@{}}

\hspace{-0.1cm}
\centering

\begin{tabular}{lccr}

\toprule
\mycol \textbf{Model} & \mycol \#Params & \mycol GFLOPs & \mycol FID \\
\midrule

SiT-B/2 & 130M & 23.0 & 33.0 \\
\arrayrulecolor{gray!50} \cmidrule(lr){1-4}
\colcyanA UDT-B/2 & \colcyanA 135M & \colcyanA 17.7 & \colcyanA 24.0 \\
\colcyanA UDT-B/2$^+$ & \colcyanA 135M & \colcyanA 17.7 & \colcyanA \textbf{20.7} \\
\arrayrulecolor{black} \bottomrule

SiT$\downarrow$-B/2 & 192M & 24.7 & 24.2 \\
SiT$\downarrow$-B/2$^{\dagger}$ & 192M & 24.7 & 20.9 \\
\arrayrulecolor{gray!50} \cmidrule(lr){1-4}
\colcyanA UDT-M/2 & \colcyanA 180M & \colcyanA 23.6 & \colcyanA 18.6 \\
\colcyanA UDT-M/2$^{\dagger}$ & \colcyanA 180M & \colcyanA 23.6 & \colcyanA \textbf{15.3} \\

\arrayrulecolor{black} \bottomrule
SiT-L/2 & 458M & 80.7 & 18.8 \\

\arrayrulecolor{gray!50} \cmidrule(lr){1-4}
\colcyanA UDT-L/2 & \colcyanA 479M & \colcyanA 61.9 & \colcyanA 9.8 \\
\colcyanA UDT-L/2$^{+}$ & \colcyanA 479M & \colcyanA 61.9 & \colcyanA \textbf{8.1} \\
\arrayrulecolor{black} \bottomrule

SiT-XL/2 & 675M & 118.6 & 17.2 \\
SiT$\downarrow$-L/2 & 679M & 81.3 & 12.8 \\
SiT$\downarrow$-L/2$^{\dagger}$ & 679M & 81.3 & 10.2 \\
\arrayrulecolor{gray!50} \cmidrule(lr){1-4}

\colcyanA UDT-XL/2 & \colcyanA 707M & \colcyanA 91.6 & \colcyanA 7.7 \\
\colcyanA UDT-XL/2$^{+}$ & \colcyanA 707M & \colcyanA 91.6 & \colcyanA \textbf{6.1} \\
\arrayrulecolor{black} \bottomrule

SiT$\downarrow$-XL/2 & 946M & 112.0 & 11.3 \\
SiT$\downarrow$-XL/2$^{\dagger}$ & 946M & 112.0 & 9.2 \\
\arrayrulecolor{gray!50} \cmidrule(lr){1-4}
\colcyanA UDT-XXL/2 & \colcyanA 935M & \colcyanA 120.5 & \colcyanA 7.0 \\
\colcyanA UDT-XXL/2$^{+}$ & \colcyanA 935M & \colcyanA 120.5 & \colcyanA \textbf{5.3} \\

\arrayrulecolor{black} \bottomrule
\end{tabular}

\end{tabular}

\end{subtable}

\end{minipage}
\hfill
\begin{minipage}{0.61\textwidth}
\caption{\textbf{System-level comparison} on ImageNet $256\!\times\!256$ with CFG. Lower $(\downarrow)$ or higher $(\uparrow)$ values are better.}
\vspace{-0.1cm}

\label{tab:system_level}
\centering

\begin{subtable}{\linewidth}
\centering
\setlength{\tabcolsep}{3.5pt}
\renewcommand{\arraystretch}{0.85}

\centering

\begin{scriptsize}

\begin{tabular}{lccccc}

\toprule
\rowcolor{violet!70!magenta!10} \textbf{Model} & \textbf{Epochs} & \textbf{Tokenizer} & \textbf{Vis. Enc.} & \textbf{FID}$\downarrow$ & \textbf{IS}$\uparrow$ \\
\midrule

\multicolumn{6}{l}{\textbf{Pixel diffusion}} \\
ADM-U~\citep{dhariwal2021beatGANs} & 400 & -- & -- & 3.94 & 186.7 \\
VDM++~\citep{kingma2023understanding} & 560 & -- & -- & \textbf{2.40} & \textbf{225.3} \\
Simple diffusion~\citep{hoogeboom2023simple} & 800 & -- & - & 2.77 & 211.8 \\
CDM~\citep{ho2022cascaded} & 2160 & -- & -- & 4.88 & 158.7 \\
\arrayrulecolor{gray} \cmidrule(lr){1-6}

\multicolumn{5}{l}{\textbf{Latent diffusion, U-Net}} \\
LDM-4~\citep{rombach2022StableDiffusion} & 200 & LDM-VAE & -- & \textbf{3.60} & \textbf{247.7} \\
\arrayrulecolor{gray} \cmidrule(lr){1-6}

\multicolumn{5}{l}{\textbf{Latent diffusion, Transformer \emph{without} representation learning}} \\
DiT-XL/2~\citep{peebles2023dit} & 1400 & SD-VAE & -- & 2.27 & 278.2 \\
SiT-XL/2~\citep{ma2024sit} & 1400 & SD-VAE & -- & 2.06 & 270.3 \\
UViT-H/2~\citep{ma2024sit} & 400 & SD-VAE & -- & 2.29 & - \\
MaskDiT~\citep{Zheng2024MaskDiT} & 1600 & SD-VAE & -- & 2.28 & 276.6 \\
DiT + TREAD~\citep{krause2025tread} & 740 & SD-VAE & -- & 1.69 & 292.7 \\
\arrayrulecolor{gray} \cmidrule(lr){1-6}
\rowcolor{blue!20!cyan!10} UDT-XL/2 (Ours) &  200 & SD-VAE & -- &  1.57 &  293.0 \\
\rowcolor{blue!20!cyan!10} UDT-XL/2 (Ours) &  500 & SD-VAE & -- &  \textbf{1.41} &  307.4 \\
\arrayrulecolor{gray!50} \cmidrule(lr){1-6}
\rowcolor{blue!20!cyan!10} UDT-XL/2$^{+}$ (Ours) &  200 & SD-VAE &  -- &  1.50 &  302.8 \\
\rowcolor{blue!20!cyan!10} UDT-XL/2$^{+}$ (Ours) &  350 & SD-VAE &  -- &  1.42 &  \textbf{308.4} \\

\arrayrulecolor{gray} \cmidrule(lr){1-6}
\multicolumn{5}{l}{\textbf{Latent diffusion, Transformer \emph{with} representation learning}} \\
SiT-XL/2 + SRA~\citep{jiang2025SREPA} & 800 & SD-VAE & -- & 1.58 & \textbf{311.4} \\
SiT-XL/2 + LSEP~\cite{yun2025_LSEP} & 800 & SD-VAE & -- & 1.46 & 296.8 \\
SiT-XL/2 + REPA~\citep{yu2025REPA} & 800 & SD-VAE & DINOv2 & 1.42 & 305.7 \\

SiT$\downarrow$-XL/2$^{\dagger}$ + UREPA~\citep{tian2026urepa} & 400 & SD-VAE & DINOv2 & 1.41 & -- \\
DDT-XL/2$^{\dagger}$ + REPA~\citep{wang2025ddt} & 400 & SD-VAE & DINOv2 & 1.40 & 303.6 \\
\arrayrulecolor{gray} \cmidrule(lr){1-6}

\rowcolor{blue!20!cyan!10} UDT-XL/2$^{+}$ + REPA (Ours) &  200 & SD-VAE &  DINOv2 &  1.44 &  296.5 \\

\rowcolor{blue!20!cyan!10} UDT-XL/2$^{+}$ + REPA (Ours) &  320 & SD-VAE &  DINOv2 &  \textbf{1.38} & 306.3 \\

\arrayrulecolor{gray} \cmidrule(lr){1-6}
\multicolumn{5}{l}{\emph{+ Improving VAE representation}} \\
DiT-XL/1$^{\dagger}$, LightningDiT~\citep{yao2025reconstruction} & 800 & VA-VAE & -- & 1.35 & 295.3 \\
\rowcolor{blue!20!cyan!10} UDT-XL/1$^{+}$ (Ours) &  500 & VA-VAE &  -- &  \textbf{1.35} & \textbf{322.0} \\
\arrayrulecolor{gray} \cmidrule(lr){1-6}
SiT-XL/1, REPA-E~\citep{leng2025REPAE} & 800 & E2E-VAE & DINOv2 & \textbf{1.26} & \textbf{314.9} \\

\arrayrulecolor{black} \bottomrule

\end{tabular}

\vspace{-0.3cm}
\end{scriptsize}

\end{subtable}

\end{minipage}
\vspace{-0.2cm}
\end{table}

We follow the experimental setup of SiT~\cite{ma2024sit} and REPA~\cite{yu2025REPA}, unless stated otherwise. All models are trained and evaluated on ImageNet~\cite{deng2009imagenet} at $256 \times 256$, following the data preprocessing protocol of ADM~\cite{dhariwal2021beatGANs}. We adopt the Base (B), Large (L), and X-Large (XL) models introduced in SiT~\cite{ma2024sit} \emph{without} modifying their configurations, as detailed in~\appref{appdx:implmentation_detail}. We implement our method in three variants. The baseline/out-of-the-box configuration, denoted as \emph{UDT}, does not include any advanced techniques and is used to verify improvements that stem purely from architectural design. The enhanced version, denoted as \emph{UDT$^+$}, incorporates advanced techniques that do not increase the number of parameters or GFLOPs, and serves as our final model. Finally, since our method is directly compatible with REPA, we also evaluate \emph{UDT$^+$ + REPA}.

\textbf{Evaluation protocol.}
We consider two complementary evaluation settings. First, we perform architectural evaluation under the standard velocity-based training objective, comparing UDT with isotropic DiT architectures such as SiT~\cite{ma2024sit} and U-Net-based transformer variants, including U-DiT~\cite{tian2024udit} and SiT$\downarrow$ from UREPA~\cite{tian2026urepa}. The goal is to elicit the improvements of our model that stem from architectural design. Second, we evaluate models augmented with REPA during training to assess how our design synergizes with advanced training strategies.

As U-DiT and SiT$\downarrow$ adopt different architectural configurations (\eg, channel dimensions or number of transformer blocks), we design additional models (S and M) to match their parameter scales for a fair comparison, as detailed in~\appref{appdx:implmentation_detail}. Optimization is performed using AdamW~\cite{kingma2015method, loshchilov2018decoupled} with learning rate $10^{-4}$. All comparisons are conducted for 80 epochs, unless stated otherwise. Detailed hyperparameters and evaluation protocol are provided in~\appref{appdx:implmentation_detail}.

\subsection{Results} \label{sec:results}

\textbf{Representation analysis.}
We evaluate representation quality via linear probing~\cite{alain2016linearprobe} on SiT-L/2, SiT$\downarrow$-L/2, and UDT-L/2, as shown in~\figref{fig:representation_vis}(b). UDT-L/2 shows higher linear probing accuracy than SiT-L/2 and reaches its peak at the bottleneck, achieving performance comparable to SiT-L/2+REPA at the target aligned layer with the visual encoder (8th layer). Complementary PCA visualizations in~\appref{appdx:pca_visualization} further demonstrate that UDT preserves more clearly structured and separable components across noise levels, indicating stronger intermediate representations.

\textbf{Architectural evaluation.} 
We first assess improvements arising purely from architectural design. \tabref{tab:FID_comparison} shows that UDT consistently outperforms isotropic DiTs (SiT) across all model scales, while significantly reducing computational cost, albeit a slight increase in parameter count due to skip connections. Notably, UDT uses $\sim$75\% of the GFLOPs of SiT through token merging while achieving substantial gains in generative performance. Compared to U-Net-based transformer variants, UDT also demonstrates a superior efficiency-performance trade-off. In particular, SiT$\downarrow$ has increased model depth, and when compared within the same model naming convention (\eg, UDT-B/2, L/2, XL/2 vs. SiT$\downarrow$-B/2, L/2, XL/2), it has substantially higher parameter counts and computational cost. Nonetheless, it is outperformed by our UDT, which uses fewer parameters and lower GFLOPs. Furthermore, for more similar parameter scales (\eg, UDT-M/2, XL/2, XXL/2 vs. SiT$\downarrow$-B/2, L/2, XL/2), the performance gap becomes even more pronounced in favor of UDT.
Finally, incorporating advanced techniques without increasing computational cost (UDT$^+$) further improves performance, again outperforming comparable enhanced 
U-Net-based transformer variants and achieving the strongest results among models without REPA. Additional comparisons with U-DiT are in~\appref{appdx:unet_dit}.

\begin{wrapfigure}[15]{r}{0.39\textwidth}
\vspace{-1.17cm}
\begin{minipage}{0.39\textwidth}
    \begin{table}[H]
\centering
\scriptsize
\setlength{\tabcolsep}{2.5pt}
\renewcommand{\arraystretch}{1.0}

\caption{FID comparison of models with REPA (w/o CFG). $\dagger, +$: using selected advanced techniques.}
\label{tab:FID_comparison_w_repa}
\vspace{+0.2cm}
\begin{tabular}{@{}ccc@{}}

\hspace{-0.1cm}
\centering

\begin{tabular}{lccr}

\toprule
\mycol \textbf{Model} & \mycol \#Params & \mycol GFLOPs & \mycol FID \\
\midrule

SiT-B/2 + REPA & 137M & 23.0 & 24.4 \\
\colcyanA UDT-B/2$^{+}$ + REPA & \colcyanA 143M & \colcyanA 17.7 & \colcyanA \textbf{16.8} \\
\arrayrulecolor{black} \bottomrule

SiT$\downarrow$-B/2$^{\dagger}$ + UREPA & 200M & 24.7 & 15.3 \\
\colcyanA UDT-M/2$^{+}$ + REPA & \colcyanA 180M & \colcyanA 23.6 & \colcyanA \textbf{12.8} \\
\arrayrulecolor{black} \bottomrule

SiT-L/2 + REPA & 466M & 82.7 & 9.7 \\
\colcyanA UDT-L/2$^{+}$ + REPA & \colcyanA 487M & \colcyanA 61.9 & \colcyanA \textbf{7.1} \\
\arrayrulecolor{black} \bottomrule

SiT-XL/2 + REPA & 683M & 118.7 & 7.9 \\
SiT$\downarrow$-L/2$^{\dagger}$ + UREPA & 687M & 81.3 & 5.8 \\ 

\colcyanA UDT-XL/2$^{+}$ + REPA & \colcyanA 715M & \colcyanA 91.6 & \colcyanA \textbf{5.7} \\

\arrayrulecolor{black} \bottomrule
SiT$\downarrow$-XL/2$^{\dagger}$ + UREPA & 954M & 109.3 & 5.4 \\
\colcyanA UDT-XXL/2$^{+}$ + REPA & \colcyanA 943M & \colcyanA 120.5 & \colcyanA \textbf{5.2} \\
\arrayrulecolor{black} \bottomrule

\end{tabular}

\end{tabular}

\end{table}
\end{minipage}
\end{wrapfigure}

\textbf{Results with REPA.} 
We next evaluate the effect of incorporating REPA for consistent model configurations. UDT and UDT$^+$ already achieve strong performance prior to alignment (\tabref{tab:FID_comparison}), in several cases matching or surpassing competing models augmented with REPA (\tabref{tab:FID_comparison_w_repa}), indicating that the architectural improvements alone provide a competitive baseline. When combined with REPA, UDT$^+$ yields further improvements that are consistently observed across all model scales. Under similar parameter scales, UDT$^+$+REPA consistently outperforms SiT+REPA and SiT$\downarrow^{\dagger}$+UREPA, with this advantage extending to larger scales where it achieves the best overall performance. Notably, these results are obtained without introducing additional architectural components or auxiliary objectives, whereas UREPA relies on extra modules and manifold-based regularization.

\textbf{System-level comparison.}
\tabref{tab:system_level} reports results using CFG~\cite{ho2021classifierfree} with a guidance interval~\cite{kynknniemi2024applying}. Our method with CFG achieves FIDs of 1.57 (UDT, out-of-the-box) and 1.50 (UDT$^+$) at only 200 epochs without explicit representation learning (middle). These results demonstrate the efficiency and strong performance of our architecture compared to the existing latent DiTs. Moreover, with continued training, UDT and UDT$^+$ \emph{without REPA} achieve FID scores of 1.41 and 1.42 at 500 and 350 epochs, respectively, outperforming the FID achieved by REPA at 800 epochs.

Combining UDT$^+$ with REPA (bottom) further improves convergence, achieving comparable generative performance to representation learning methods in fewer epochs and ultimately reaching a SOTA FID of 1.38 at 320 epochs under SD-VAE-f8d4.  Lastly, with an improved VAE (\ie, VA-VAE-f16d32~\cite{yao2025reconstruction}), our UDT$^+$ achieves an FID of 1.35 at 500 epochs, matching the FID achieved by LightningDiT after 800 epochs while attaining a higher IS. Detailed evaluations are provided in~\appref{appdx:FID_details}. We also note that recent works have adopted class-balanced generation~\cite{leng2025REPAE, zheng2026RAE, li2025jit, wang2025ddt}, which generally improves FID scores. Comparisons under this setting are provided in~\appref{appdx:fid_class_balance}. Representative qualitative samples are provided in~\appref{appdx:qual_results}.

\begin{figure*}[!t]
    \centering
    \includegraphics[width=.98\linewidth]{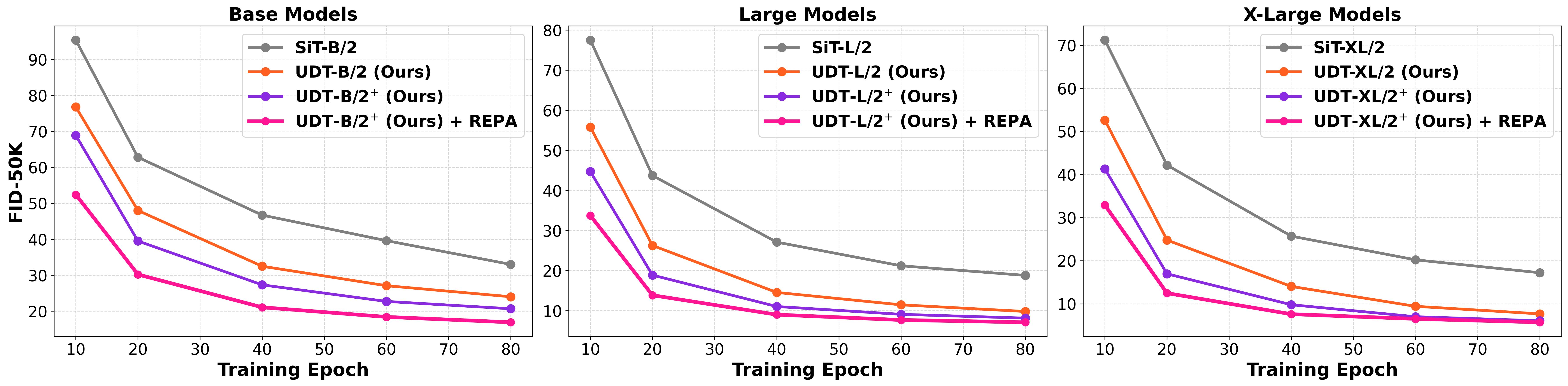} 
    \caption{FID vs. Epoch on ImageNet $256 \times 256$ without CFG.}
    \vspace{-3ex}
    \label{fig:fid_epoch}
\end{figure*}

\textbf{Training efficiency (FID vs. epoch).}
Our proposed method achieves faster convergence, measured by FID over training epochs, as shown in \figref{fig:overview}(b) and ~\figref{fig:fid_epoch}. \figref{fig:fid_epoch} shows that UDT achieves the FID attained by the baseline SiT model at 80 epochs, in under 40 epochs across all model sizes. With architectural optimization, UDT$^+$, and UDT$^+$+REPA achieve the same FID in under 30 and 20 epochs, respectively. For longer training (\figref{fig:overview}(b)), UDT-XL/2 and its variants achieve $20$--$40\times$ faster convergence compared to SiT-XL/2 models, while also consistently outperforming SiT-XL/2 + REPA.
Further analyses of computational efficiency are provided in~\appref{appdx:training_efficiency}.

\subsection{Design Choices and Extensions} \label{sec:ablations}
\textbf{Token merge strategies.}
We follow the strategy of~\cite{bolya2022tome}, computing 
key-based similarity, tracking the number of merged tokens, and progressively merging tokens using weighted averaging as in~\eqnref{eq:tome_merge}. Proportional attention is also applied to preserve consistent attention contributions after merging. \tabref{tab:ablation_tome}(c) presents the ablation results with and without each component, showing that this 
design achieves the best performance in diffusion transformer training. 

\textbf{Ablation study on advanced techniques.}
We adopt log-normal time sampling~\cite{esser2024mmdit} with $(\mu=0.0, \sigma=1.0)$, along with RoPE~\cite{su2024roformer} and SwiGLU~\cite{shazeer2020glu}, and perform ablations on ImageNet $256 \times 256$ using UDT-B/2 trained for 80 epochs. As shown in~\tabref{tab:ablation_advanced_tech}, each component improves performance by 0.7--2.7 FID, and when combined, they synergistically yield a total improvement of 3.3 FID. Therefore, we adopt all components in our final model, denoted as UDT$^+$.

\begin{table}[t]
\centering

\begin{minipage}{0.27\textwidth}
\centering
\caption{Ablation results for different advanced components on UDT-B/2.}
\vspace{+0.2cm}
\label{tab:ablation_advanced_tech}

\begin{subtable}{\linewidth}
\centering
    \scriptsize
    \centering
    \setlength{\tabcolsep}{1.0pt}
    \renewcommand{\arraystretch}{1.15}
      
\begin{tabular}{@{}cccc@{}}
\toprule
\mycol Log-normal & \mycol SwiGLU & \mycol RoPE & \mycol FID \\
\midrule
\xmark  &  \xmark &  \xmark & 24.0 \\  
\cmarkB &         &         & 21.3 \\  
        & \cmarkB &         & 23.3 \\
        &         & \cmarkB & 22.8 \\
\cmarkB & \cmarkB & \cmarkB & \textbf{20.7} \\
\bottomrule
\end{tabular}

\end{subtable}
\vspace{+0.2cm}
\end{minipage}
\hfill
\begin{minipage}{0.43\textwidth}
\centering
\caption{Computational comparison between patch sizes 1 and 2. Ratios (\textcolor{red}{red}, \textcolor{blue}{blue}) are relative to SiT models with patch size 2.}
\label{tab:fid_patch1}
\vspace{-0.2cm}

\begin{subtable}{\linewidth}
\centering
\centering
\tiny
\setlength{\tabcolsep}{0.1pt}
\renewcommand{\arraystretch}{0.7}

\begin{tabular}{@{}lcc|lccc@{}}
\toprule
\mycol Model & \mycol\#Params & \mycol GFLOPs & \mycol Model & \mycol \#Params & \mycol GFLOPs & \mycol FID \\
\midrule

SiT-B/2 & 130M & 23.0 & UDT-B/2$^+$ & 135M & 17.7 & 20.7 \\
SiT-B/1 & 131M & 106.4 {\textcolor{red}{(\tiny 4.6$\times$)}}  & \colcyanA UDT-B/1$^+$ & \colcyanA 136M & \colcyanA 49.8 {\textcolor{blue}{(\tiny 2.2$\times$)}} & \colcyanA \textbf{16.6} \\

\arrayrulecolor{gray!50}
\cmidrule(lr){1-7}

SiT-L/2 & 458M & 80.7 & UDT-L/2$^+$ & 479M & 61.9 & 8.1 \\
SiT-L/1 & 459M & 361.0 {\textcolor{red}{(\tiny 4.5$\times$)}} & \colcyanA UDT-L/1$^+$ & \colcyanA 480M & \colcyanA 116.1 {\textcolor{blue}{(\tiny 1.4$\times$)}} & \colcyanA \textbf{7.6} \\

\arrayrulecolor{gray!50}
\cmidrule(lr){1-7}

SiT-XL/2 & 675M & 118.6  & UDT-XL/2$^+$ & 707M & 91.6 & 6.1 \\
SiT-XL/1 & 676M & 524.6 {\textcolor{red}{(\tiny 4.4$\times$)}}  & \colcyanA UDT-XL/1$^+$ & \colcyanA 708M & \colcyanA 158.1 {\textcolor{blue}{(\tiny 1.3$\times$)}} & \colcyanA \textbf{5.8} \\

\arrayrulecolor{black}
\bottomrule
\end{tabular}

\end{subtable}

\end{minipage}
\hfill
\begin{minipage}{0.27\textwidth}
\centering
\caption{Comparison on ImageNet $512\!\times\!512$ with CFG and interval.}
\vspace{-0.2cm}
\label{tab:fid_512}

\begin{subtable}{\linewidth}
\centering
    \tiny
    \centering
    \renewcommand{\arraystretch}{0.9}
    \setlength{\tabcolsep}{0.5pt}

\begin{tabular}{@{}lccc@{}}
\toprule
\mycol Model & \mycol  Epoch & \mycol GFLOPs & \mycol FID \\
\midrule
DiT-XL/2  &  600 & 524.6 & 3.04 \\  
SiT-XL/2  &  600 & 524.6 & 2.62 \\  
SiT-XL/2 + REPA  & 200 & 524.6 & 2.10 \\  
\arrayrulecolor{gray!50}
\cmidrule(lr){1-4}
\colcyanA UDT-XL/2$^+$ & \colcyanA 100 & \colcyanA 158.1 & \colcyanA 2.00 \\ 
\colcyanA UDT-XL/2$^+$  & \colcyanA \textbf{340} & \colcyanA \textbf{158.1} & \colcyanA \textbf{1.71} \\ 
\arrayrulecolor{gray!50}
\cmidrule(lr){1-4}
\colcyanA UDT-XL/2$^+$ (FT)  & \colcyanA \textbf{200} & \colcyanA \textbf{158.1} & \colcyanA \textbf{1.58} \\ 
\arrayrulecolor{black}
\bottomrule

\end{tabular}

\end{subtable}

\end{minipage}
\vspace{-0.5cm}
\end{table}


\textbf{Extending to longer token sequences.} 
Latent-space DiTs typically use a patch size of 2, balancing performance and efficiency~\cite{ma2024sit,peebles2023dit,yu2025REPA,tian2026urepa,esser2024mmdit}. A patch size of 1 provides finer spatial detail but significantly increases sequence length and cost~\cite{ma2024sit,peebles2023dit}. Our UDT naturally extends to such longer-sequence settings by applying high rates of token reduction in early encoder blocks, followed by progressive merging/unmerging. For example, with patch size 1 on $32\!\times\!32$ latent input (1024 tokens), early merging (e.g., 50\%) quickly reduces the sequence to the standard 256-token regime, without substantial degradation due to similarity among such small patches. The model then proceeds with standard progressive reduction toward $N_{\textrm{Merge}}$, enabling efficient training and inference while retaining hierarchical U-Net modeling.

\tabref{tab:fid_patch1} (left) shows that the original SiT exhibits increased computational cost when patch size is reduced from 2 to 1, consistently increasing by $\sim$4.5$\times$. In contrast, our method 
increases computation by only 1.3--2.2$\times$, and this ratio decreases as the model size increases. As a result, our approach efficiently handles longer sequences (\eg, 1024 instead of 256) while improving FID performance. Due to the substantial computational cost of running SiT with patch size = 1, no FID results were provided in~\cite{ma2024sit}, thus FID results are only reported for our model UDT$^+$.

Additional experiments on ImageNet $512\times512$ with a patch size of 2 further support this observation. As shown in~\tabref{tab:fid_512}, UDT$^+$, without REPA, achieves an FID of 2.00 at 100 epochs, outperforming existing DiT/SiT models as well as REPA at 200 epochs (FID 2.10). Ultimately, UDT$^+$ achieves an FID of 1.71 when trained from scratch and 1.58 when fine-tuned from a mid-training checkpoint at a resolution of $256 \times 256$, while using only 30\% of the GFLOPs required by isotropic DiTs. Additional comparisons and implementation details for ImageNet $512\times512$ are provided in~\appref{appdx:fid_512}.

\textbf{Drop-in Replacement for Isotropic DiTs.}
Our method can be broadly applied as a drop-in replacement for isotropic transformer architectures to improve both training efficiency and generative performance. As representative examples, we consider DiT~\cite{peebles2023dit} with $\epsilon$ prediction, pixel-space DiTs such as JiT~\cite{li2025jit}, models using improved VAE architectures such as VA-VAE~\cite{yao2025reconstruction}, as previously discussed in the system-level comparison in \secref{sec:results}, and T2I models such as MMDiT~\cite{esser2024mmdit}. For the first three architectures, we replace the DiT blocks with our UDT. For MMDiT, which comprises separate text and visual embedding branches, we replace the isotropic DiT-based visual branch with our UDT, denoted as MMUDT.

\tabref{tab:udt_drop_in} demonstrates that replacing DiT with our UDT architecture consistently improves both training efficiency and generative performance: (a) UDT, out-of-the-box, with $\epsilon$ prediction outperforms both DiT and DiT+REPA. (b) UDT is also applicable to pixel-level DiTs, achieving improved FID scores over JiT both with and without CFG. (c) Combined with the improved VA-VAE, our method demonstrates faster convergence, achieving an FID of 1.35 in 500 epochs, matching the FID achieved by LightningDiT after 800 epochs. (d) In T2I generation on MSCOCO~\cite{lin2014mscoco}, our method further improves generative quality. These results demonstrate the \emph{broad applicability of our method} as a drop-in replacement for existing isotropic transformer models. Detailed implementations and results are provided in~\appref{appdx:UDT_DiT_variants}.

\begin{wrapfigure}[9]{r}{0.37\textwidth}
\vspace{-1.1cm}
\begin{minipage}{0.37\textwidth}
    \begin{table}[H]
\centering
\tiny
\setlength{\tabcolsep}{3.5pt}
\renewcommand{\arraystretch}{0.5}

\caption{Comparison on Full and 10\% ImageNet Subsets without CFG.}
\label{tab:FID_subset}
\vspace{+0.1cm}
\begin{tabular}{llccc}
    \toprule
    \mycol  \textbf{Dataset} & \mycol  \textbf{Model} & \mycol  \textbf{Epoch} & \mycol  \textbf{Total Images} & \mycol  \textbf{FID} \\
    \midrule
    Full set & SiT-L/2 & 80 & 102.5M & 18.8 \\
    \midrule
    \multirow{8}{*}{\makecell{Subset\\(10\%)}} & SiT-L/2 & 500 & 64.1M & 22.9 \\
    \arrayrulecolor{gray!50} \cmidrule{2-5}
    & \colcyanA UDT-L/2 & \colcyanA 300 & \colcyanA 38.4M & \colcyanA 13.7 \\
    & \colcyanA & \colcyanA 500 & \colcyanA \textbf{64.1M} & \colcyanA \textbf{10.6} \\
    \arrayrulecolor{gray!50} \cmidrule{2-5}
    & \colcyanA UDT-L/2$^+$  & \colcyanA 200 & \colcyanA 25.6M & \colcyanA 12.7 \\
    &\colcyanA & \colcyanA 300 & \colcyanA \textbf{38.4M} & \colcyanA \textbf{10.3} \\
    \arrayrulecolor{black}  \bottomrule
\end{tabular}

\end{table}
\end{minipage}
\end{wrapfigure}

\textbf{Efficacy on Reduced Training Data.}
In many real-world applications, such as medical imaging, large-scale datasets are often unavailable~\cite{litjens2017survey,yun2025_TE-MRI_NIPS,alcalar2025_CUPID}. To evaluate UDT under limited data regimes, we train on only 10\% of ImageNet while retaining all 1,000 classes, reducing the training set from 1.28M to 128K images. As shown in Table~\ref{tab:FID_subset}, we compare UDT against SiT-L/2 without CFG. With the limited dataset, SiT-L/2 achieves an FID of 22.9 at 500 epochs, underperforming its full-dataset counterpart (FID 18.8 at 80 epochs). In contrast, UDT-L/2 achieves FIDs of 13.7 and 10.6 at 300 and 500 epochs, respectively, while UDT$^+$-L/2 further accelerates convergence, reaching FIDs of 12.7 at 200 epochs and 10.3 at 300 epochs, surpassing SiT-L/2 trained with the full database at 80 epochs. Notably, despite achieving an even better FID, UDT$^+$-L/2 trained on the subset dataset (300 epochs) requires $\sim 2.67\times$ less wall-clock training time than SiT-L/2 trained on the full dataset (80 epochs), as it processes substantially fewer training images. These results demonstrate that UDT provides substantially improved data efficiency and faster convergence compared to isotropic DiT architectures, highlighting its effectiveness in limited-data generation scenarios. 

\begin{table*}[!t]
\centering
\tiny
\caption{UDT as a Drop-in Replacement for DiT Variants. Comparison on ImageNet (a,b,c) and MSCOCO (d) at $256\!\times\!256$. ${\dagger}, +$: using their selected architectural optimizations. *: with CFG.}
\vspace{-2ex}
\label{tab:udt_drop_in}
\vspace{+0.2cm}
\setlength{\tabcolsep}{2pt}

\resizebox{\textwidth}{!}{%
\begin{tabular}{cccc}

\renewcommand{\arraystretch}{0.9}
\begin{tabular}{lcc}
\toprule
\mycol Model & \mycol Epoch & \mycol FID$\downarrow$ \\
\midrule
DiT-B/2 & 80 & 43.5 \\
\colcyanA UDT-DiT-B/2 & \colcyanA 80 & \colcyanA \textbf{30.6} \\
\arrayrulecolor{gray!50}
\cmidrule(lr){1-3}

DiT-L/2 & 80 & 23.3 \\
DiT-L/2 + REPA & 80 & 15.6 \\
\colcyanA UDT-DiT-L/2 & \colcyanA 80 & \colcyanA \textbf{13.5} \\
\arrayrulecolor{gray!50}
\cmidrule(lr){1-3}

DiT-XL/2 & 80 & 19.5 \\
DiT-XL/2 + REPA & 80 & 12.3 \\
\colcyanA UDT-DiT-XL/2 & \colcyanA 80 & \colcyanA \textbf{11.2} \\

\arrayrulecolor{black}
\bottomrule
\end{tabular}
\vspace{+0.1cm}

&

\renewcommand{\arraystretch}{0.6}
\begin{tabular}{lccc}
\toprule
\mycol Model & \mycol Epoch & \mycol FID$\downarrow$ & \mycol FID$^*$$\downarrow$ \\
\midrule
JiT-B/16$^\dagger$ & 50 & 82.6 & 13.6 \\
JiT-B/16$^\dagger$ & 200 & 66.1 & 4.7 \\

\colcyanA UDT-JiT-B/16$^+$ & \colcyanA 100 & \colcyanA 64.6 & \colcyanA 6.3 \\
\colcyanA UDT-JiT-B/16$^+$ & \colcyanA 200 & \colcyanA \textbf{58.9} & \colcyanA \textbf{4.6} \\
\arrayrulecolor{gray!50} \cmidrule(lr){1-4}

JiT-L/16$^\dagger$ & 50  & 62.2 & 6.9 \\
JiT-L/16$^\dagger$ & 200  & 48.0 & 3.0 \\
\colcyanA UDT-JiT-L/16$^+$  & \colcyanA 50 & \colcyanA 47.9 & \colcyanA 5.8 \\
\colcyanA UDT-JiT-L/16$^+$  & \colcyanA 200 & \colcyanA \textbf{37.7} & \colcyanA \textbf{2.8} \\

\arrayrulecolor{black}
\bottomrule
\end{tabular}

&

\renewcommand{\arraystretch}{1.45}
\begin{tabular}{lcccc}
\toprule
\mycol Model & \mycol Tokenizer & \mycol Epoch & \mycol FID$^*$$\downarrow$ & \mycol IS$^*$$\uparrow$ \\
\midrule
LightningDiT-XL/1$^\dagger$  & VA-VAE & 64 & 2.11 & 252.3 \\
\colcyanA UDT-XL/1$^+$  & \colcyanA VA-VAE & \colcyanA 64 & \colcyanA  \colcyanA \textbf{1.99} & \colcyanA \textbf{265.8} \\
\arrayrulecolor{gray!50}
\cmidrule(lr){1-5}
LightningDiT-XL/1$^\dagger$ & VA-VAE & 800 &  1.35 & 295.3 \\
\colcyanA UDT-XL/1$^+$  & \colcyanA VA-VAE & \colcyanA \textbf{500} & \colcyanA \textbf{1.35} & \colcyanA \textbf{322.0} \\
\arrayrulecolor{black}
\bottomrule
\end{tabular}

&

\renewcommand{\arraystretch}{1.4}
\begin{tabular}{lrc}
\toprule
\mycol Model & \mycol Iter. & \mycol FID$^*$$\downarrow$ \\
\midrule
MMDiT & 150K & 5.5 \\
\colcyanA MMUDT & \colcyanA 150K & \colcyanA \textbf{4.7} \\

\arrayrulecolor{black}
\bottomrule
\end{tabular}

\\

\tiny (a) $\epsilon$-prediction (DiT~\cite{peebles2023dit})
&
\tiny (b) Pixel-Level DiT (JiT~\cite{li2025jit})
&
\tiny (c) VAE Modification (VA-VAE~\cite{yao2025reconstruction})
&
\tiny (d) T2I Model (MMDiT~\cite{esser2024mmdit})

\end{tabular}%
}
\vspace{-3.5ex}
\end{table*}

\section{Conclusion}

We present the U-Net Diffusion Transformer (UDT), which combines the representation power of DiTs with the architectural advantages of encoder--decoder U-Nets. Our key idea is data-adaptive token merging for transformer-specific down/upsampling, facilitating hierarchical representation learning while improving computational efficiency. UDT significantly accelerates training convergence and achieves strong generative performance out-of-the-box. Its variants further improve performance with architectural refinements and REPA, reaching an FID of 1.38 on XL models after only 320 epochs with SD-VAE, and further improves to 1.35 after 500 epochs with the stronger VA-VAE. Finally, UDT serves as a drop-in replacement for a broad range of DiT variants, including pixel-level DiTs and T2I models such as MMDiT, and works efficiently with longer token sequences, \eg, at a resolution of 512$\times$512.

\bibliographystyle{abbrv}
\small
\bibliography{refs_base}
\normalsize
\newpage
\appendix

{\Large \textbf{Appendix}}\par

\section{Implementation Details} \label{appdx:implmentation_detail}

\subsection{Model Configurations}
All trainings were conducted from scratch, using 4 NVIDIA A100 GPUs for the 80-epoch setup and 8 NVIDIA A100 GPUs for higher epoch training setups.

\begin{table}[h]
\centering
\caption{Hyperparameter setup for different UDT model size with patch size 2.}
\vspace{+0.2cm}
\renewcommand{\arraystretch}{1.05}
\begin{small}

\resizebox{\textwidth}{!}{%
\begin{tabular}{lcccccc}
\toprule
 & \textbf{UDT-S/2} & \textbf{UDT-B/2} & \textbf{UDT-M/2} & \textbf{UDT-L/2} & \textbf{UDT-XL/2} & \textbf{UDT-XXL/2}  \\
\midrule
\textbf{Architecture} & & & \\
Input dim. & $32\times32\times4$ & $32\times32\times4$ & $32\times32\times4$ & $32\times32\times4$ & $32\times32\times4$ & $32\times32\times4$ \\
Num. layers (Enc.--Dec.) & 12 (6--6) & 12 (6--6) & 16 (8--8) & 24 (12--12) & 28 (14--14) & 30 (15--15) \\
Hidden dim. & 480 & 768 & 768 & 1,024 & 1,152 & 1,280 \\
Num. heads & 8 & 12 & 12 & 16 & 16 & 16 \\
\midrule
\textbf{Token Merge} & & & \\
$N_\textrm{Merge} $  & 112 & 112 & 112 & 112 & 112 & 112 \\
$r$ schedule & 36 (Enc 2--5) & 36 (Enc 2--5) & 24 (Enc 2--7) & 15 (Enc 2-5), & 12 (Enc 2--13) & 12 (Enc 2), \\
 &  & & & 14(Enc 6-11) & & 11(Enc 3-14) \\

\midrule
\textbf{+REPA} (if used) & & & \\
Align. Depth & -- & 6 & 8 & 12 & 14 & 15 \\
Weight for REPA loss & -- & 0.5 & 0.5 & 0.5 & 0.5 & 0.5 \\

\midrule
\textbf{Optimization} & & & \\

Batch size & 256 & 256 & 256 & 256 & 512 & 512  \\
Optimizer & AdamW & AdamW & AdamW & AdamW & AdamW & AdamW \\
lr & 0.0001 & 0.0001 & 0.0001 & 0.0001 & 0.0001  & 0.0001 \\
$(\beta_1,\beta_2)$ & (0.9,0.999) & (0.9,0.999) & (0.9,0.999) & (0.9,0.999) & (0.9,0.999) & (0.9,0.999) \\
\midrule
\textbf{Interpolants} & & & & & \\
$\alpha_t$ & $1-t$ & $1-t$ & $1-t$ & $1-t$ & $1-t$ & $1-t$ \\
$\sigma_t$ & $t$ & $t$ & $t$ & $t$ & $t$ & $t$ \\
$w_t$ & $\sigma_t$ & $\sigma_t$ & $\sigma_t$ & $\sigma_t$ & $\sigma_t$ & $\sigma_t$ \\
Training objective & v-prediction & v-prediction & v-prediction & v-prediction & v-prediction & v-prediction \\
Sampler & Euler-Maruyama & Euler-Maruyama & Euler-Maruyama & Euler-Maruyama & Euler-Maruyama & Euler-Maruyama \\
Sampling steps & 250 & 250 & 250 & 250 & 250 & 250 \\

\bottomrule
\label{tab:appdx_hyperparameters}
\end{tabular}
}
\end{small}
\end{table}

\begin{table}[h]
\centering
\caption{Hyperparameter setup for different UDT model size with patch size 1.}
\vspace{+0.2cm}
\setlength{\tabcolsep}{3.0pt} 
\renewcommand{\arraystretch}{0.9}
\begin{footnotesize}
\begin{tabular}{lcccc}
\toprule
 & \textbf{UDT-B/1} & \textbf{UDT-L/1} & \textbf{UDT-XL/1}\\
\midrule
\textbf{Architecture} & & & \\
Input dim. & $32\times32\times4$ & $32\times32\times4$ & $32\times32\times4$  \\
Num. layers (Enc.--Dec.) & 12 (6--6) & 24 (12--12) & 28 (14--14) \\
Hidden dim. & 768 & 1,024 & 1,152 \\
Num. heads & 12 & 16 & 16 \\
\midrule
\textbf{Token Merge} & & & \\
$N_\textrm{Merge} $  & 112 & 112 & 112 \\
$r$ schedule & 512 (Enc 2), & 512 (Enc 2),  & 512 (Enc 2), \\
 & 256 (Enc 3),  & 256 (Enc 3), & 256 (Enc 3),\\
 & 72 (Enc 4-5)  & 18 (Enc 4-11) & 15 (Enc 4-17), 14(Enc 8-13)\\

\midrule
\textbf{Optimization} & & & \\
Training epochs & 80 & 80 & 80\\
Batch size & 256 & 256 & 512 \\
Optimizer & AdamW & AdamW & AdamW \\
lr & 0.0001 & 0.0001 & 0.0001 \\
$(\beta_1,\beta_2)$ & (0.9,0.999) & (0.9,0.999) & (0.9,0.999) \\
\midrule
\textbf{Interpolants} & & &  \\
$\alpha_t$ & $1-t$ & $1-t$ & $1-t$ \\
$\sigma_t$ & $t$ & $t$ & $t$ \\
$w_t$ & $\sigma_t$ & $\sigma_t$ & $\sigma_t$ \\
Training objective & v-prediction & v-prediction & v-prediction  \\
Sampler & Euler-Maruyama & Euler-Maruyama  & Euler-Maruyama  \\
Sampling steps & 250 & 250 & 250  \\

\bottomrule
\label{tab:appdx_hyperparameters_patch1}
\end{tabular}

\end{footnotesize}
\end{table}

\subsection{Evaluation Protocol}
We follow the ADM evaluation protocol~\cite{dhariwal2021beatGANs} for assessing image generation quality. We report standard metrics including Fréchet Inception Distance (FID)~\cite{heusel2017gans} and Inception Score (IS)~\cite{salimans2016improved}, computed over 50K generated samples. Following SiT~\cite{ma2024sit} and REPA~\cite{yu2025REPA}, we use the SDE-based Euler–Maruyama sampler with 250 steps. Evaluation is conducted on 50K ImageNet~\cite{deng2009imagenet} validation images resized to $256 \times 256$ or $512 \times 512$ using 4 A100-40GB GPUs with seed = 0.

\subsection{Implementation of Linear Probing}
Linear probing accuracy is evaluated across different depths for SiT-L/2, SiT-L/2+REPA, SiT$\downarrow$-L/2 from UREPA, and UDT-L/2 (ours). Specifically, we follow the experimental setups of~\cite{xiang2023DDAE, yu2025REPA} with minor modifications. For each pretrained model, we perform linear probing by attaching a batch normalization layer followed by a linear classifier, trained on the ImageNet training set for 90 epochs with a batch size of 6144. We use the Adam optimizer with a cosine decay learning rate scheduler, with an initial learning rate of 0.001. Evaluation is conducted on the 50K ImageNet validation set.


\begin{table}[!b]
\centering
\setlength{\tabcolsep}{7.0pt}
\renewcommand{\arraystretch}{1.0}

\caption{FID of UDT variants across training epochs with CFG weights and guidance intervals on ImageNet $256 \times 256$. \colorbox{lime!35}{\makebox[0.4cm][l]{\rule{0pt}{1.4ex}}} and
\colorbox{green!25}{\makebox[0.4cm][l]{\rule{0pt}{1.4ex}}} indicate performance surpassing the FID achieved by SiT-XL at 1400 epochs and REPA at 800 epochs, respectively. The \textbf{best} results for each model are bolded.}

\vspace{+0.2cm}
\label{tab:UDT_cfg}
\begin{scriptsize}
\begin{tabular}{lc|cc|cccccc}
\toprule
\rowcolor{violet!70!magenta!10}
Model & Epoch. & $w_\text{cfg}$ & Interval & FID$\downarrow$ & sFID$\downarrow$ & IS$\uparrow$ & Pre.$\uparrow$ & Rec.$\uparrow$ \\
\midrule
\mycolB SiT-XL/2 (Baseline)~\cite{ma2024sit}  & \mycolC 1400 & 1.5 & [0, 1.0]  & \mycolC 2.06 & 4.50 & 270.3 & 0.82 & 0.59 \\
\mycolB SiT-XL/2 + REPA~\cite{yu2025REPA} & \mycolD 800 & 1.8 & [0, 0.7]  & \mycolD 1.42 & 4.70 & 305.7 & 0.80 & 0.64 \\
\midrule
\multicolumn{9}{l}{\textbf{UDT}} \\
UDT-XL/2 (ours) & \mycolC 100 & 1.7 & [0, 0.7] & \mycolC 1.95 & 4.56 & 266.6 & 0.82 & 0.59 \\
UDT-XL/2 (ours) & 200 & 1.7 & [0, 0.7] & 1.57 & 4.35 & 293.0 & 0.81 & 0.63 \\
UDT-XL/2 (ours) & 300 & 1.7 & [0, 0.7] & 1.48 & 4.33 & 299.3 & 0.80 & 0.63 \\
UDT-XL/2 (ours) & 400 & 1.7 & [0, 0.7] & 1.44 & 4.29 & 305.7 & 0.80 & 0.64 \\
UDT-XL/2 (ours) & \mycolD 410 & 1.7 & [0, 0.7] & \mycolD 1.42 & 4.30 & 306.1 & 0.80 & 0.65 \\
UDT-XL/2 (ours) & \textbf{500} & 1.7 & [0, 0.7] & \textbf{1.41} & \textbf{4.28} & \textbf{307.4} & \textbf{0.80} & \textbf{0.65} \\

\midrule
\multicolumn{9}{l}{\textbf{UDT$^+$}} \\
UDT$^+$-XL/2 (ours) & \mycolC 80 & 1.7 & [0, 0.7] & \mycolC 1.99 & 4.71 & 276.5 & 0.81 & 0.60 \\
UDT$^+$-XL/2 (ours) & 100 & 1.7 & [0, 0.7] & 1.82 & 4.50 & 287.2 & 0.82 & 0.60 \\
UDT$^+$-XL/2 (ours) & 200 & 1.7 & [0, 0.7] & 1.51 & 4.29 & 302.8 & 0.80 & 0.64 \\
UDT$^+$-XL/2 (ours) & 300 & 1.7 & [0, 0.7] & 1.49 & 4.37 & 308.1 & 0.80 & 0.64 \\
UDT$^+$-XL/2 (ours) & \mycolD 320 & 1.7 & [0, 0.7] & \mycolD 1.42 & 4.36 & 305.4 & 0.80 & 0.64 \\
UDT$^+$-XL/2 (ours) & \textbf{350} & 1.7 & [0, 0.7] & \textbf{1.42} & \textbf{4.32} & \textbf{308.4} & \textbf{0.80} & \textbf{0.65} \\
\midrule
\multicolumn{9}{l}{\textbf{UDT$^+$ + REPA}} \\
UDT$^+$-XL/2 + REPA (ours) & \mycolC 50 & 1.7 & [0, 0.7] & \mycolC 1.88 & 4.54 & 260.3 & 0.79 & 0.62 \\
UDT$^+$-XL/2 + REPA (ours) & 100 & 1.7 & [0, 0.7] & 1.55 & 4.33 & 283.9 & 0.79 & 0.64 \\
UDT$^+$-XL/2 + REPA (ours) & 200 & 1.7 & [0, 0.7] & 1.44 & 4.35 & 296.5 & 0.79 & 0.65 \\
UDT$^+$-XL/2 + REPA (ours) & \mycolD 250 & 1.7 & [0, 0.7] & \mycolD 1.42 & 4.47 & 298.2 & 0.79 & 0.65 \\
UDT$^+$-XL/2 + REPA (ours) & 280 & 1.7 & [0, 0.7] & 1.40 & 4.38 & 304.4 & 0.79 & 0.65 \\
UDT$^+$-XL/2 + REPA (ours) & \textbf{320} & 1.7 & [0, 0.7] & \textbf{1.38} & \textbf{4.38} & \textbf{306.3} & \textbf{0.79} & \textbf{0.66} \\

\bottomrule
\end{tabular}
\end{scriptsize}
\label{tab:appdx_FID_all_udt}
\end{table}

\begin{table}[!t]
\centering
\setlength{\tabcolsep}{7.0pt}
\renewcommand{\arraystretch}{1.0}

\caption{FID of UDT$^+$ across training epochs with CFG weights ($\omega_{CFG}=2.2$ and guidance intervals [0, 0.75]) on ImageNet $512 \times 512$. 
\colorbox{green!25}{\makebox[0.4cm][l]{\rule{0pt}{1.4ex}}} indicates performance surpassing the FID achieved by REPA at 200 epochs, respectively. The \textbf{best} results for each model are bolded.}

\vspace{+0.2cm}
\label{tab:UDT_cfg_512}
\begin{scriptsize}
\begin{tabular}{lc|cccccc}
\toprule
\rowcolor{violet!70!magenta!10}
Model & Epoch. & FID$\downarrow$ & sFID$\downarrow$ & IS$\uparrow$ & Pre.$\uparrow$ & Rec.$\uparrow$ \\
\midrule
\mycolB DiT-XL/2~\citep{peebles2023dit} & 600 & 3.04 & 5.02 & 240.8 & 0.84 & 0.54 \\
\mycolB SiT-XL/2~\citep{ma2024sit} & 600 & 2.62 & 4.18 & 252.2 & 0.84 & 0.57 \\
\mycolB MaskDiT~\citep{Zheng2024MaskDiT} & 800 & 2.50 & 5.10 & 256.3 & 0.84 & 0.57 \\
\mycolB SiT-XL/2 + REPA~\citep{yu2025REPA} & \mycolD 200 & \mycolD 2.08 & 4.19 & 274.6 & 0.83 & 0.58 \\

\midrule
\multicolumn{7}{l}{\textbf{UDT$^+$}} \\
UDT$^+$-XL/2 (ours) & \mycolD 100  & \mycolD 2.00 & 4.86 & 288.0 & 0.80 & 0.62 \\
UDT$^+$-XL/2 (ours) & 150  & 1.85 & 4.81 & 311.0 & 0.79 & 0.65 \\
UDT$^+$-XL/2 (ours) & 200  & 1.79 & 4.83 & 304.6 & 0.79 & \textbf{0.65} \\
UDT$^+$-XL/2 (ours) & 300  & 1.76 & 4.75 & \textbf{317.5} & 0.80 & 0.62 \\
UDT$^+$-XL/2 (ours) & \textbf{340}  & \textbf{1.71} & \textbf{4.67} & 316.2 & \textbf{0.80} & 0.64 \\
\midrule
\multicolumn{7}{l}{\textbf{UDT$^+$ (Fine-tuning)}} \\
UDT$^+$-XL/2 (ours) & \mycolD 30  & \mycolD 1.82 & 4.81 & 320.5 & 0.79 & 0.63 \\
UDT$^+$-XL/2 (ours) & 100  & 1.63 & 4.68 & 319.1 & 0.79 & 0.64 \\
UDT$^+$-XL/2 (ours) & \textbf{200}  & \textbf{1.58} & \textbf{4.58} & \textbf{327.7} & \textbf{0.79} & \textbf{0.64} \\

\bottomrule
\end{tabular}
\end{scriptsize}
\label{tab:appdx_FID_all_udt_512}
\end{table}

\section{Detailed Evaluation of UDT Variants} \label{appdx:FID_details}

\subsection{Experiments on ImageNet 256$\times$256}

We report detailed generation results over epochs for each UDT model with CFG on ImageNet $256\times256$, as shown in~\tabref{tab:appdx_FID_all_udt}. UDT-XL/2 (out-of-the-box), UDT$^+$-XL/2, and UDT$^+$-XL/2 with REPA achieve SiT-XL/2's 2.06 FID at 1400 epochs in under 100, 80, and 50 epochs, respectively. Furthermore, all models surpass SiT-XL/2 + REPA's 1.42 FID at 800 epochs in under 410, 350, and 250 epochs, respectively. Notably, UDT-XL/2, an out-of-the-box model \emph{without} REPA or architectural optimization, achieves an FID of 1.41 at 500 epochs. 

\subsection{Experiments on ImageNet $512 \times 512$} \label{appdx:fid_512}
We trained UDT$^+$-XL/2 both from scratch, and also by adopting the fine-tuning strategy of DDT~\cite{wang2025ddt} that initializes from a mid-training checkpoint obtained at a resolution of $256 \times 256$. For the latter, we initialized from the 250-epoch checkpoint of the UDT$^+$-XL/2 model trained at $256 \times 256$ and fine-tuned it for an additional 200 epochs.

As shown in~\tabref{tab:UDT_cfg_512}, UDT$^+$-XL/2 \emph{without} REPA achieves an FID of 2.00 after only 100 epochs of training from scratch, outperforming existing DiT-based models as well as REPA, while requiring substantially fewer training epochs. It further improves to an FID of 1.76 after 300 epochs. Moreover, by adopting the fine-tuning strategy, UDT$^+$-XL/2 achieves a SOTA FID of 1.58 after only 200 fine-tuning epochs on ImageNet $512 \times 512$.

\section{Comparison with U-Net DiTs} \label{appdx:unet_dit}

We compare two U-Net DiT variants, including U-DiT~\cite{tian2024udit} and SiT$\downarrow$ from UREPA~\cite{tian2026urepa}. Where necessary, we re-implement the baselines with and without additional techniques, marking the enhanced variants with $\dagger$. SiT$\downarrow$ variants are compared in~\secref{sec:results}. In this section, we provide comparison results with U-DiT~\cite{tian2024udit}.

\textbf{Comparison with U-DiT.}
U-DiT is another U-Net-based transformer architecture. A direct comparison requires additional adjustments, as U-DiT fixes the number of transformer blocks to 22, performs two stages of downsampling and upsampling, and follows the standard U-Net design by increasing and decreasing channel dimensions across stages. Moreover, its model scaling differs from the conventional DiT naming scheme. 

\begin{wrapfigure}[15]{r}{0.41\textwidth} 
\vspace{-1.2cm} 
\begin{minipage}{0.41\textwidth} 
\begin{table}[H]
\scriptsize
\centering
\setlength{\tabcolsep}{7.5pt}
\renewcommand{\arraystretch}{0.9}
\captionof{table}{Comparison with U-DiTs on ImageNet $256\!\times\!256$ (80 epochs) w/o CFG. 
}
\label{tab:udit_comparison}
\vspace{0.2cm}

\begin{tabular}{lccc}
\toprule
\mycol Model & \mycol \#Params & \mycol GFLOPs & \mycol FID \\
\midrule

U-DiT-S & 52M & 5.9 & 41.0 \\
U-DiT-S$^{\dagger}$ & 59M & 6.0 & \textbf{31.5} \\
\arrayrulecolor{gray!50} \cmidrule(lr){1-4}
\colcyanA UDT-S/2 & \colcyanA 53M & \colcyanA 6.6 & \colcyanA \textbf{39.3} \\
\colcyanA UDT-S/2$^+$ & \colcyanA 53M & \colcyanA 6.6 & \colcyanA 34.6 \\
\arrayrulecolor{black} \midrule

U-DiT-B & 204M & 22.0 & 20.9 \\
U-DiT-B$^{\dagger}$ & 231M & 22.2 & 16.7 \\
\arrayrulecolor{gray!50} \cmidrule(lr){1-4}
\colcyanA UDT-M/2 & \colcyanA 180M & \colcyanA 23.6 & \colcyanA \textbf{18.6} \\
\colcyanA UDT-M/2$^+$ & \colcyanA 180M & \colcyanA 23.6 & \colcyanA \textbf{15.3} \\
\arrayrulecolor{black} \midrule

U-DiT-L & 810M & 84.5 & 12.0 \\
U-DiT-L$^{\dagger}$ & 916M & 85.0 & 10.1 \\
\arrayrulecolor{gray!50} \cmidrule(lr){1-4}
\colcyanA UDT-XL/2 & \colcyanA 707M & \colcyanA 91.6 & \colcyanA \textbf{7.7} \\
\colcyanA UDT-XL/2$^+$ & \colcyanA 707M & \colcyanA 91.6 & \colcyanA \textbf{6.1} \\
\arrayrulecolor{black}

\bottomrule
\end{tabular}
\end{table}
 
\end{minipage} 
\end{wrapfigure}

To ensure a fair comparison, we redesign our models to match the parameter count and GFLOPs of U-DiT-S and U-DiT-B, resulting in UDT-S/2 and UDT-M/2, respectively, as detailed in~\appref{appdx:implmentation_detail}. For U-DiT-L, we perform a direct comparison with UDT-XL/2. All our counterparts maintain fewer parameters than U-DiT.

As illustrated in~\tabref{tab:udit_comparison}, the results show that our method outperforms U-DiT in all cases except for U-DiT-S with advanced techniques (vs. UDT-S/2), and the performance gap becomes more pronounced as the model size increases. We note that U-DiT uses full tokens at the input and output stages (\ie, 1024 tokens for $32 \times 32$ latent input) without patchification, and applies two stages of $2\times$ spatial downsampling, resulting in a sharp reduction in token length (1024 $\rightarrow$ 256 $\rightarrow$ 64). Despite operating with patch size 2, our method achieves better performance than U-DiT, even though U-DiT processes full tokens in the latent space. Furthermore, we note that our method can be also be used with patch size 1, yielding additional improvements, as discussed in ~\secref{sec:ablations}.

\section{Drop-in Replacement for Isotropic DiT Variants} \label{appdx:UDT_DiT_variants}

UDT serves as a drop-in replacement for a wide range of isotropic DiT variants, including DiT~\cite{peebles2023dit} with $\epsilon$ prediction, pixel-space DiTs such as JiT~\cite{li2025jit}, models using modified VAE architectures such as VA-VAE~\cite{yao2025reconstruction}, and text-to-image models such as MMDiT~\cite{esser2024mmdit}. Unless otherwise stated, we follow the original training and sampling protocols of each baseline and replace only the backbone architecture with UDT. This enables a fair comparison and isolates the performance gains attributable to the proposed architecture.

\subsection{Applying UDT to $\epsilon$-prediction DiT}  \label{appdx:udt_dit}

\begin{wrapfigure}[14]{r}{0.42\textwidth} 
\vspace{-1.9cm} 
\begin{minipage}{0.42\textwidth} 
\begin{table}[H]
\scriptsize
\centering
\setlength{\tabcolsep}{6.3pt}
\renewcommand{\arraystretch}{1.1}
\captionof{table}{UDT-DiT ($\epsilon$-prediction) on ImageNet $256 \times 256$ (80 epochs) w/o CFG. 
}
\vspace{0.2cm}
\label{tab:udt_dit}

\begin{tabular}{lccc}
\toprule
\mycol Model & \mycol \#Params & \mycol GFLOPs & \mycol FID \\
\midrule

DiT-B/2 & 130M & 23.0 & 43.5 \\
\arrayrulecolor{gray!50} \cmidrule(lr){1-4}
\colcyanA UDT-DiT-B/2 & \colcyanA 135M & \colcyanA 17.7 & \colcyanA \textbf{30.6} \\
\colcyanA UDT-DiT-B/2$^+$ & \colcyanA 135M & \colcyanA 17.7 & \colcyanA \textbf{29.1} \\
\arrayrulecolor{black} \bottomrule

DiT-L/2 & 458M & 80.7 & 23.3 \\
DiT-L/2 + REPA & 458M & 80.7 & 15.6 \\
\arrayrulecolor{gray!50} \cmidrule(lr){1-4}
\colcyanA UDT-DiT-L/2 & \colcyanA 479M & \colcyanA 61.9 & \colcyanA \textbf{13.5} \\
\colcyanA UDT-DiT-L/2$^{+}$ & \colcyanA 479M & \colcyanA 61.9 & \colcyanA \textbf{12.4} \\
\arrayrulecolor{black} \bottomrule

DiT-XL/2 & 675M & 118.6 & 19.5 \\
DiT-XL/2 + REPA & 675M & 118.6 & 12.3 \\
\arrayrulecolor{gray!50} \cmidrule(lr){1-4}
\colcyanA UDT-DiT-XL/2 & \colcyanA 707M & \colcyanA 91.6 & \colcyanA \textbf{11.2} \\
\colcyanA UDT-DiT-XL/2$^{+}$ & \colcyanA 707M & \colcyanA 91.6 & \colcyanA \textbf{9.6} \\
\arrayrulecolor{black}

\bottomrule
\end{tabular}
\end{table}

\end{minipage} 
\end{wrapfigure}

In the main text, UDT models are trained with velocity prediction following the protocol of SiT~\cite{ma2024sit}. UDT can be extended to different objective functions such as $\epsilon$-prediction, as in DiT~\cite{peebles2023dit}. Accordingly, we replace the isotropic DiT architecture with our proposed UDT and conduct experiments with both $\epsilon$ and variance prediction settings, denoting this variant as UDT-DiT for clarity.

As shown in~\tabref{tab:udt_dit}, UDT-DiT significantly improves generative performance across model scales compared to DiT. Notably, even the out-of-the-box UDT-DiT models outperform DiT+REPA. The UDT-DiT$^+$ variant further improves performance. For the $^+$ model, we additionally apply SwiGLU and partial RoPE.

\subsection{Applying UDT to Pixel-Level DiT (JiT)} \label{appdx:udt_jit}

Our UDT can also be applied to pixel-level DiTs. JiT~\cite{li2025jit} revisits pixel-space diffusion and shows that plain ViT architectures with larger patch sizes can be trained end-to-end on raw images without a latent tokenizer. Since ViT-based models share the same transformer blocks as DiT/SiT~\cite{peebles2023dit, ma2024sit}, differing mainly in the embedding of noisy inputs, they can be naturally replaced with our U-shaped UDT. We use the official JiT implementation for training and evaluation.

JiT improves performance by introducing a bottleneck embedding and adopting several architectural enhancements, including in-context class conditioning (appending 32 class tokens), SwiGLU, RMSNorm, RoPE, QK-Norm, and lognormal sampling. We integrate our UDT$^+$ into JiT, denoted as UDT$^+$-JiT, adopting all architectural enhancements except RMSNorm. For in-context class conditioning, JiT appends 32 additional class tokens from a specific layer onward (\eg, the 4th layer in the Base model and the 8th layer in the Large model) and maintains them until the final layer. In contrast, UDT applies in-context class tokens only around the bottleneck stages, specifically to 3 layers in the Base model and 6 layers in the Large model. This design reduces computational overhead while applying conditioning to layers with enhanced representations. 

We train both the Base and Large models from scratch for 200 epochs, using a global batch size of 512 and a learning rate of $1.5\times10^{-4}$ for the Base model, and a global batch size of 256 with a learning rate of $1\times10^{-4}$ for the Large model. All other training configurations follow those of JiT. For sampling, we also follow the JiT's Euler 50-step sampling protocol with CFG and a CFG interval. We set $\omega_{\mathrm{CFG}} = 3.0$ and $2.5$ for the Base and Large models, respectively, with a guidance interval of $[0.1, 1.0]$. Note that JiT uses the opposite noise schedule convention from ours, where 0 corresponds to pure noise and 1 corresponds to a clean sample.

\subsection{Improving UDT with a Stronger VAE (VA-VAE)} \label{appdx:udt-va-vae}

LightningDiT~\cite{yao2025reconstruction} utilizes a vision foundation model-aligned variational autoencoder (VA-VAE). Specifically, it follows a two-stage training pipeline: first, SD-VAE-f16d32 is fine-tuned using REPA with pre-trained ViTs such as DINOv2, resulting in VA-VAE-f16d32, which achieves strong performance in both generation and reconstruction. This improved VAE encoder generates latent features $\mathbf{z} \in \mathbb{R}^{16 \times 16 \times 32}$ for $256\times256$ image resolution. Using these latent features, a DiT model with a patch size of 1 is trained with additional architectural optimizations and a velocity direction loss~\cite{yao2024fasterdit}, further improving training effectiveness. 

\begin{wrapfigure}[11]{r}{0.45\textwidth} 
\vspace{-1.2cm} 
\begin{minipage}{0.45\textwidth} 
\begin{table}[H]
\scriptsize
\centering
\setlength{\tabcolsep}{2.5pt}
\renewcommand{\arraystretch}{1.1}
\captionof{table}{UDT$^+$ with VA-VAE on ImageNet $256 \times 256$. *: results with CFG.
}
\vspace{0.2cm}
\label{tab:udt_vavae}

\begin{tabular}{lc|cc|cc}
\toprule
\mycol Model & \mycol Epoch & \mycol FID$\downarrow$ & \mycol IS$\uparrow$ & \mycol FID$^*$$\downarrow$ & \mycol IS$^*$$\uparrow$ \\
\midrule

LightningDiT-XL/1 & 64 & 5.14 & 130.2 & 2.11 & 252.3 \\
\colcyanA UDT-XL/1$^+$ & \colcyanA 64 & \colcyanA 4.90 & \colcyanA 145.8 & \colcyanA 1.99 & \colcyanA 265.8 \\
\arrayrulecolor{gray!50} \cmidrule(lr){1-6}

\colcyanA UDT-XL/1$^+$ & \colcyanA 200 & \colcyanA 3.21 & \colcyanA 182.6 & \colcyanA 1.54 & \colcyanA 313.6 \\
\arrayrulecolor{gray!50} \cmidrule(lr){1-6}

\colcyanA UDT-XL/1$^+$ & \colcyanA \textbf{500} & \colcyanA 2.45 & \colcyanA \textbf{205.7} & \colcyanA \textbf{1.35} & \colcyanA \textbf{322.0} \\

\arrayrulecolor{gray!50} \cmidrule(lr){1-6}

LightningDiT-XL/1 & 800 & \textbf{2.17} & 205.6 & 1.35 & 295.3 \\

\arrayrulecolor{black}

\bottomrule
\end{tabular}
\end{table}
 
\end{minipage} 
\end{wrapfigure}

We replace the DiT backbone in LightningDiT-XL/1 with UDT$^+$-XL/1, adopting the same architectural optimizations and training configuration, except that we use QK-Norm instead of RMSNorm for training stability and set the global batch size to 512 and the learning rate to $1.5\times10^{-4}$, instead of the batch size of 1024 and learning rate of $2\times10^{-4}$ used in LightningDiT. For sampling, we follow the LightningDiT protocol, specifically using Euler sampling with 250 steps and a timestep shift (shift factor = 0.3). We set $\omega_{\mathrm{CFG}} = 2.8$ and the guidance interval to $[0.3, 1.0]$. Note that LightningDiT uses the opposite noise schedule convention from ours, where 0 corresponds to pure noise and 1 corresponds to a clean sample.

As shown in Table~\ref{tab:udt_vavae}, UDT demonstrates faster convergence, achieving an FID of 1.99 after only 64 epochs, surpassing LightningDiT, which reaches an FID of 2.11 with CFG. UDT further achieves an FID of 1.35 at 500 epochs, matching LightningDiT's 800-epoch performance, while obtaining a higher IS score.

\subsection{Applying UDT to T2I model}  \label{appdx:t2i_detail}
We conduct T2I experiments on the MSCOCO dataset~\cite{lin2014mscoco}, following the protocol in~\cite{bao2023uvit, yu2025REPA}. We replace the DiT backbone in MMDiT with UDT, denoted as MMUDT. Specifically, we set both MMDiT and MMUDT to 24 layers with a hidden dimension of 768, following REPA~\cite{yu2025REPA}. The former consists of an isotropic DiT backbone, while the latter uses our proposed UDT. We train MMDiT and MMUDT from scratch for 150K iterations with a batch size of 256 on the MSCOCO $256\times256$. We use the token merging configuration of the Large model for MMUDT, as described in~\appref{appdx:implmentation_detail}. For sampling, we use an SDE-based sampler with 250 steps and follow prior work by setting $\mathrm{CFG}=2$~\cite{bao2023uvit, yu2025REPA}.

\section{Analysis of Computational Efficiency} 
\label{appdx:training_efficiency}

\subsection{Training Efficiency: Wall-Clock Time vs. FID}

\begin{figure*}[h]
    \centering
    \includegraphics[width=1.0\linewidth]{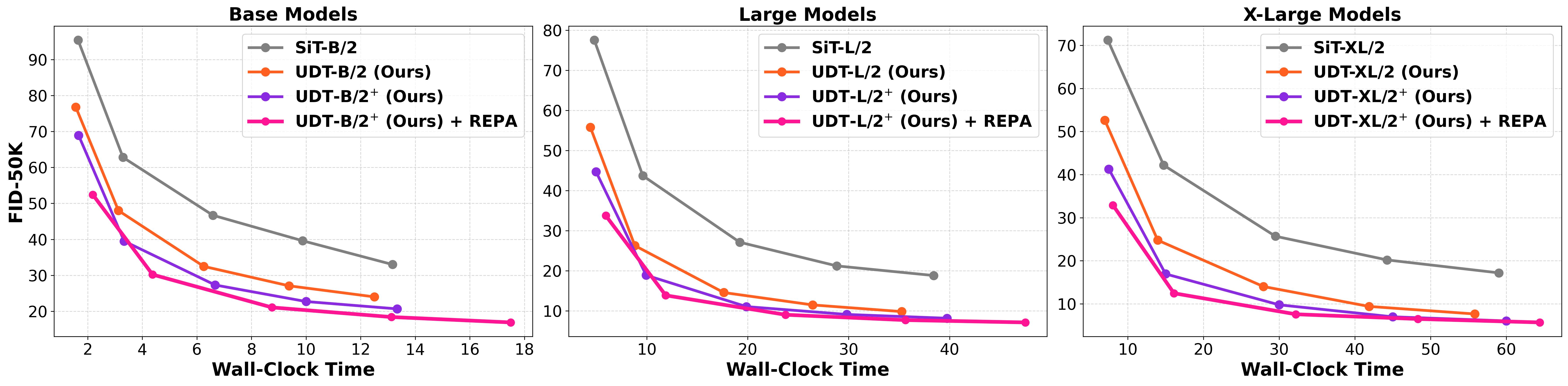} 
    \vspace{-0.1cm}
    \caption{FID vs. Wall-Clock Time (hours) on ImageNet  $256 \times 256$ without CFG. }
    \label{fig:fid_wct}
\end{figure*}

We analyze the training efficiency of our model and its variants using Wall-Clock Time (WCT) vs. FID, compared to SiT with measurements taken at $\{10, 20, 40, 60, 80\}$ epochs. UDT achieves faster convergence and reaches the final performance at 80 epochs more efficiently than SiT, thanks to reduced computational cost enabled by token merging-based data-adaptive downsampling, while also showing faster FID convergence.

UDT$^{\dagger}$ slightly slows down training due to additional architectural components (e.g., SwiGLU and RoPE), but it remains comparable to SiT in terms of WCT, while achieving better FID vs. time trade-offs. Finally, UDT$^{\dagger}$+REPA introduces additional computational overhead from extracting features from a visual encoder and applying a regularization loss (\ie, patch-level alignment), resulting in longer WCT to reach 80 epochs. Nonetheless, it achieves the fastest FID convergence WCT due to the benefits of representation alignment with strong ViT representations.

\subsection{Analysis of the Computational Overhead of Token Merging/Unmerging.}

\begin{table}[!h]
    \scriptsize
    \centering
    \setlength{\tabcolsep}{5.0pt}
    \renewcommand{\arraystretch}{1.0}
    \caption{Comparison of computational efficiency among SiT-L/2, SiT$\downarrow$-L/2, and UDT-L/2.
    }
    \vspace{+0.1cm}
    \label{tab:compu_efficiency}
    \begin{tabular}{lcccccc}
    \toprule
    Method & Att. Time (ms) & MLP (ms) & down/upsampling (ms) & skip connection (ms) & Total Time (ms) & GPU Memory (GB) \\
    \midrule
    \multicolumn{7}{l}{\textbf{Training}} \\
    SiT-L/2 & 207.34 & 368.85 & -- & -- & 1836.89 & 35.66 \\
    SiT$\downarrow$-L/2 & 185.10 & 333.24 & 3.50 & 35.13 & 1783.41 & 36.13 \\
    UDT-L/2 & 148.42 & 260.91 & 9.58 & 29.67 & 1486.18 & 29.97 \\
    \midrule 
    
    \multicolumn{7}{l}{\textbf{Inference}} \\
    SiT-L/2 & 207.27 & 368.83 & -- & -- & 596.68 & 4.18 \\
    SiT$\downarrow$-L/2 & 184.66 & 333.08 & 3.49 & 35.12 & 576.69 & 6.33 \\
    UDT-L/2 & 148.34 & 260.85 & 9.25 & 29.65 & 476.67 & 4.50 \\
    \arrayrulecolor{black} \bottomrule
        
    \end{tabular}
\end{table}

While UDT introduces additional operations for token merging and unmerging, the overhead is negligible compared to the computational savings from reducing the number of tokens processed by the transformer blocks. To quantitatively analyze the overhead, we measured the training and inference time of the Large models on 4 NVIDIA A100-40GB GPUs with a global batch size of 256 and averaged the results over 30 iterations. 

As shown in~\tabref{tab:compu_efficiency}, both SiT$\downarrow$ and UDT introduce additional costs during both training and inference due to the downsampling/upsampling operations. Specifically, SiT$\downarrow$ uses fixed $2\times2$ spatial token downsampling and upsampling, while UDT performs adaptive token merging and unmerging. In addition, both U-shaped architectures introduce overhead from skip connections with linear projections. However, these additional costs are outweighed by the computational savings from reducing the number of tokens processed by the transformer blocks. Compared to SiT, token reduction decreases the computation of the attention and MLP blocks by 89\% for SiT$\downarrow$ and 70\% for UDT.

Consequently, after accounting for both the overhead of the downsampling/upsampling operations and the computational savings in the attention and MLP blocks for training, UDT-L/2 requires only 81\% of the per-batch training time and 84\% of the GPU memory of SiT-L/2, whereas SiT$\downarrow$ exhibits a similar training time but slightly higher GPU memory consumption. The reported training time includes both the forward and backward passes. 
For inference, UDT-L/2 requires only 80\% of the per-batch inference time while using a comparable amount of GPU memory to SiT-L/2, whereas SiT$\downarrow$ exhibits a similar inference time but requires 1.5$\times$ more GPU memory.

A key difference between SiT$\downarrow$ and UDT is that SiT$\downarrow$ increases the number of transformer blocks from 24 to 32 to match the GLOPs of SiT-L/2. This results in a significant increase in the number of parameters (from 458M to 679M) and prevents it from achieving meaningful computational savings in terms of training/inference time and GPU memory. In contrast, UDT introduces only additional parameters from the skip connections (479M parameters in total) while achieving faster training and inference with reduced GPU memory usage during training.

These results demonstrate that UDT achieves improved computational efficiency in terms of both wall-clock time and GPU memory usage while maintaining strong generative performance.

\section{Representation Analysis}
\label{appdx:pca_visualization}

\subsection{PCA Visualization}

\begin{figure*}[h]
    \centering
    \includegraphics[width=1.0\linewidth]{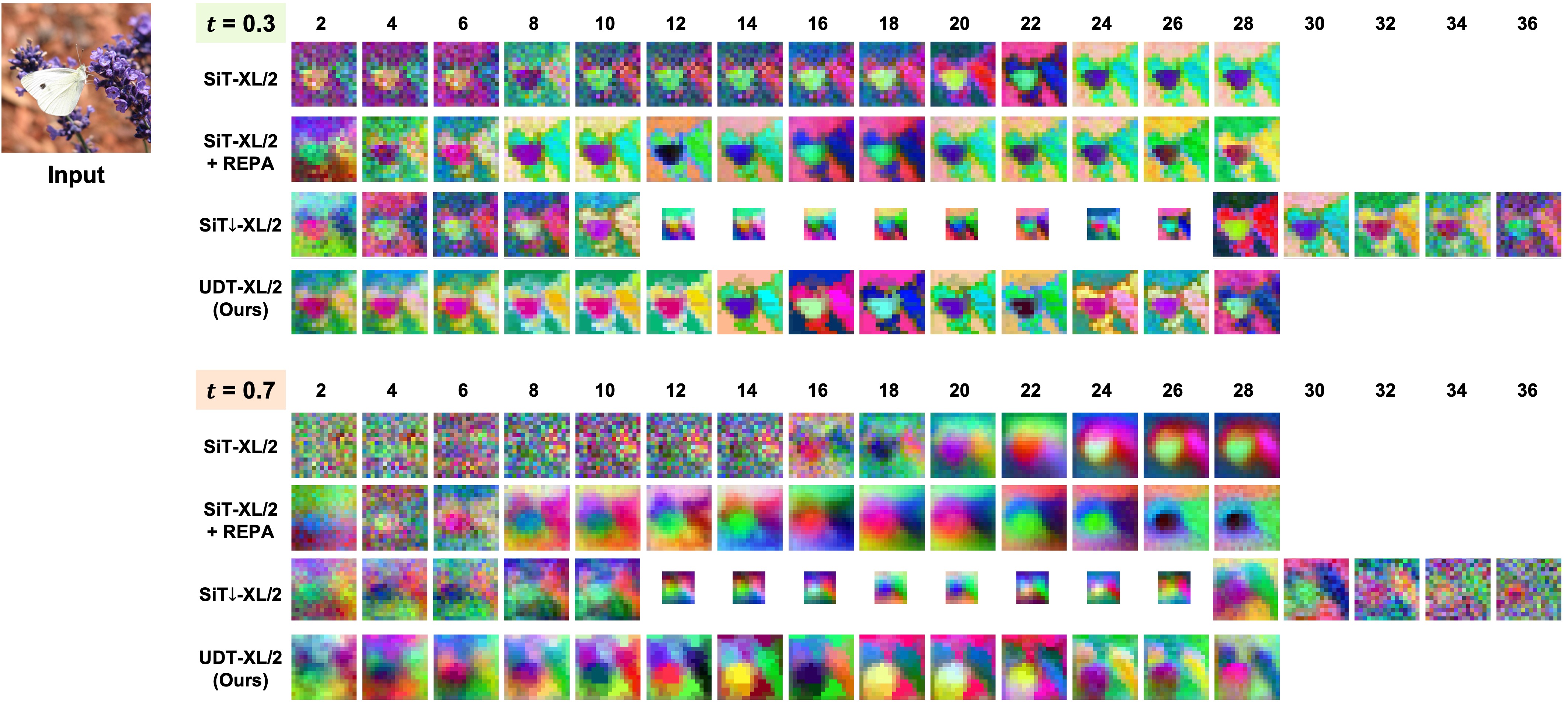} 
    \vspace{-0.1cm}
    \caption{PCA visualization of intermediate features at $t \in \{0.3, 0.7\}$ for XL/2 models without advanced techniques, trained for 80 epochs (zoom in for a more detailed view). SiT$\downarrow$-XL/2 has 36 layers in total, while the others have 28 layers.}
    \label{fig:pca_appdx}
    
\end{figure*}

We visualize the PCA of intermediate features from each trained XL/2 model as shown in~\figref{fig:pca_appdx}. The SiT-XL/2 exhibits noisy representations in the early layers at $t=0.3$, and it shows little semantic structure until the mid-layers under higher noise levels (e.g., $t=0.7$). SiT-XL/2+REPA, which targets the 8th layer for representation learning, produces clearer semantic structures at both $t=0.3$ and $t=0.7$ compared to SiT-XL/2.

SiT$\downarrow$-XL/2, a U-shape diffusion transformer, shows improved representations over the SiT-XL/2 in the early layers, however, its later representations become noisy, and spatial reduction (to $H/2, W/2$) leads to blurred fine-grained details. In contrast, our UDT-XL/2 produces sharper and more clearly separated representations across all depths, consistently outperforming other models under both low- and high-noise conditions.

\subsection{Downstream Evaluation of UDT Representations on Semantic Segmentation}

\begin{wrapfigure}[8]{r}{0.30\textwidth} 
\vspace{-1.2cm} 
\begin{minipage}{0.30\textwidth} 
 \begin{table}[H]
\centering
\tiny
\setlength{\tabcolsep}{3.5pt}
\renewcommand{\arraystretch}{1.0}
\caption{Comparison of linear semantic segmentation performance on PASCAL data.}
\vspace{+0.1cm}
\label{tab:rep_semantic}
\begin{tabular}{llccc}
    \toprule
      \textbf{Method} & Layer (Total) & \textbf{mIoU} & \textbf{pixAcc} \\
    \midrule
    SiT-L/2 & 16 (24) & 38.30 & 84.81$\%$\\
    \midrule
    SiT$\downarrow$-L/2 & 24 (32)  & 41.02 & 85.92$\%$\\
    UDT-L/2 & 16 (24) & \textbf{55.00} & \textbf{89.78}$\%$\\
    \arrayrulecolor{black}  \bottomrule
\end{tabular}

\end{table}

\end{minipage} 
\end{wrapfigure}

To evaluate whether UDT representations transfer well to dense prediction tasks, we conducted semantic segmentation experiments on the PASCAL VOC 2012 dataset~\cite{everingham2010pascal} (train: 1,464 images; val: 1,449 images). We followed the DINOv2~\cite{oquab2024dinov} protocol for linear semantic segmentation. Specifically, each pretrained Large backbone (80 epochs) was frozen. Input images were resized to $256\times256$, encoded into the latent space, noised at $t=0.1$, and passed through the frozen backbone to obtain patch-token representations. Only a linear segmentation head was trained on top of these extracted features. The segmentation head predicts a $16\times16$ logit map, which is bilinearly upsampled to $256\times256$ for loss computation and evaluation.

For each model, we extracted features from the layer that achieved the highest linear classification probing accuracy, as shown in Fig. 2(b). For UDT, we extracted features from the 16th layer and applied token unmerging to recover the original $16\times16$ token grid. For SiT and UREPA, we extracted features from the 16th and 24th layers, respectively, which already have a $16\times16$ token grid. For training, we used a global batch size of 256, a learning rate of $5\times10^{-3}$, the AdamW optimizer, and trained the linear head for 100 epochs. We evaluate performance using mIoU and pixel accuracy.

As shown in~\tabref{tab:rep_semantic}, the representations from UDT-L/2 achieve the best segmentation performance in terms of both mIoU and pixel accuracy. These dense prediction results are consistent with the linear probing results on the classification task in the initial submission, further demonstrating that our proposed method effectively learns transferable and semantically meaningful representations.

\clearpage

\section{Evaluation with class-balanced 50K sampling.}
\label{appdx:fid_class_balance}

\begin{wrapfigure}[8]{l}{0.55\textwidth} 
\vspace{-0.5cm} 
\begin{minipage}{0.55\textwidth} 
\setlength{\tabcolsep}{10.0pt}
\renewcommand{\arraystretch}{1.0}

\centering
\caption{Evaluation with class-balanced 50K sampling on ImageNet $256 \times 256$. \textcolor{black!50}{Gray: results with auto-guidance instead of CFG.}}
\label{tab:fid_class_balance}

\begin{tiny}

\begin{tabular}{lccc}

\toprule
\rowcolor{violet!70!magenta!10} \textbf{Model} & \textbf{Epochs} & \textbf{Vis. Enc.} & \textbf{FID}$\downarrow$  \\
\midrule

SiT-XL/2~\citep{ma2024sit} & 1400 & -- & 1.95 \\
SiT-XL/2 + REPA~\citep{yu2025REPA} & 800 & DINOv2 & 1.29 \\

DDT-XL/2$^{\dagger}$ + REPA~\citep{wang2025ddt} & 400 & DINOv2 & 1.26 \\

\rowcolor{blue!20!cyan!10} UDT-XL/2$^{+}$ + REPA (Ours) &  \textbf{320} &  DINOv2 &  \textbf{1.26} \\

\arrayrulecolor{gray} \cmidrule(lr){1-4}
\multicolumn{4}{l}{\emph{with improved VAE}} \\
SiT-XL/1, REPA-E~\citep{leng2025REPAE} & 800 & DINOv2 & 1.15 \\

\textcolor{black!50}{DiT$^{\textrm{DH}}$-XL/158, RAE~\citep{leng2025REPAE}} & \textcolor{black!50}{800} & \textcolor{black!50}{DINOv2} & \textcolor{black!50}{1.13} \\

\arrayrulecolor{black} \bottomrule

\end{tabular}

\end{tiny}

\end{minipage} 
\end{wrapfigure}

Recent studies have adopted class-balanced generation \cite{leng2025REPAE, zheng2026RAE, li2025jit}, which generally results in improved FID scores. To ensure a fair comparison, we evaluate our method under the same setting and compare it with the reimplemented results reported in RAE~\cite{zheng2026RAE}. We use CFG weights ($\omega_{\text{CFG}} = 1.71$) and guidance interval $[0, 0.69]$. 

\vspace{1cm}

\section{Additional Analyses.}
\subsection{Effects of Each Component (Token Merging and Skip Connections).}

\begin{table}[H]
\centering
\scriptsize
\setlength{\tabcolsep}{3.5pt}
\renewcommand{\arraystretch}{1.0}
\caption{Ablation study of token merging/unmerging and skip connections.}
\label{tab:FID_component}
\vspace{+0.1cm}
\begin{tabular}{llccc}
    \toprule
      \textbf{Method} & \textbf{Epoch} & \textbf{FID} \\
    \midrule
    SiT-B/2 & 80 & 33.0 \\
    \midrule
    (a) SiT-B/2 + Token Merge/Unmerge  & 80 & 293.0 \\
    (b) SiT-B/2 + Skip Connection & 80 & 31.5 \\
    (c) SiT-B/2 + Token Merge/Unmerge + Skip Connection (UDT-B/2) & 80 & 24.0 \\
    \arrayrulecolor{black}  \bottomrule
\end{tabular}

\end{table}

To better isolate each component's contribution, we conducted additional experiments on an isotropic DiT backbone (SiT-B/2): (a) adding token merging/unmerging for downsampling/upsampling, and (b) adding skip connections without any spatial token reduction.

As shown in~\tabref{tab:FID_component}, token merging reduces the spatial resolution and inevitably discards some high-frequency information. Without skip connections, the model cannot effectively recover this information, resulting in severely degraded generative performance. This highlights the necessity of skip connections when introducing token merging. On the other hand, adding skip connections alone to the isotropic SiT results in a U-shaped architecture similar to U-ViT. Note this leads to a simpler architecture than prior U-shaped DiTs (e.g., U-ViT, U-DiT, and SiT$\downarrow$), thus the benefit from skip connections alone ends up being limited.

Finally, (c), our full UDT-B/2, combines token merging/unmerging with skip connections. The two components work synergistically, enabling efficient hierarchical token processing while preserving high-frequency information through skip connections, resulting in the best generative performance. 

\subsection{Impact of Different Downsampling Methods.}

\begin{wrapfigure}[8]{r}{0.35\textwidth} 
\vspace{-1.2cm} 
\begin{minipage}{0.35\textwidth} 
\begin{table}[H]
\centering
\scriptsize
\setlength{\tabcolsep}{3.5pt}
\renewcommand{\arraystretch}{1.0}
\caption{Comparison of downsampling strategies for UDT-B/2 based on FID.}
\label{tab:FID_downsampling}
\vspace{+0.1cm}
\begin{tabular}{llccc}
    \toprule
      \textbf{Method} & Downsampling & \textbf{Epoch} & \textbf{FID} \\
    \midrule
    SiT-B/2 & - & 80 & 33.0 \\
    UDT-B/2 & Avg. pooling  & 80 & 29.2 \\
    UDT-B/2 & Strided Conv. & 80 & 32.3 \\
    UDT-B/2 & Token Merge & 80 & 24.0 \\
    \arrayrulecolor{black}  \bottomrule
\end{tabular}

\end{table}
\end{minipage} 
\end{wrapfigure}

Existing U-shaped DiTs, such as U-DiT and SiT$\downarrow$, already investigate fixed spatial downsampling strategies (e.g., convolution-based downsampling and patch merging) within the DiT architecture. Our work directly replaces these fixed downsampling operations with adaptive token merging, while keeping the overall encoder--decoder design comparable. Therefore, the comparisons with U-DiT and SiT$\downarrow$ can also be viewed as a comparison between fixed spatial downsampling and adaptive token merging under similar architectural settings.

Furthermore, we conducted additional experiments with several representative downsampling strategies. We considered average pooling and strided convolution as alternatives to token merging. Specifically, average pooling used a $2\times2$ kernel with stride 2, while strided convolution used a $3\times3$ kernel with stride 2. Both methods reduced the spatial token resolution by $2\times$ while preserving the token embedding dimension. We evaluated these approaches on UDT-B/2 by replacing token merging with each alternative. Among the 12 transformer blocks, spatial token reduction was applied only to the middle 4 blocks, while the first and last 4 blocks maintained full token resolution with skip connections. As shown in~\tabref{tab:FID_downsampling}, both downsampling strategies improve the generative performance over SiT-B/2, with average pooling outperforming strided convolution. However, our proposed token merging approach consistently achieves the best FID, demonstrating that adaptive token reduction is more effective than fixed local downsampling methods.

\section{Limitations \& Impact}  \label{appdx:limitation_impact}
\textbf{Limitations.} The scope of this paper does not cover advanced generative settings such as video generation and 2K higher-resolution images. Further investigation is warranted to evaluate the applicability and scalability of our framework in these scenarios.

\textbf{Broader Impact and Safeguards.} As this work focuses on AI-generated content (AIGC), there is a possibility that the outputs may include inappropriate material. It is therefore important to remain mindful of the potential negative societal implications.

\clearpage
\section{Qualitative Results}  \label{appdx:qual_results}

\begin{figure*}[h]
    \centering
    \includegraphics[width=0.85\linewidth]{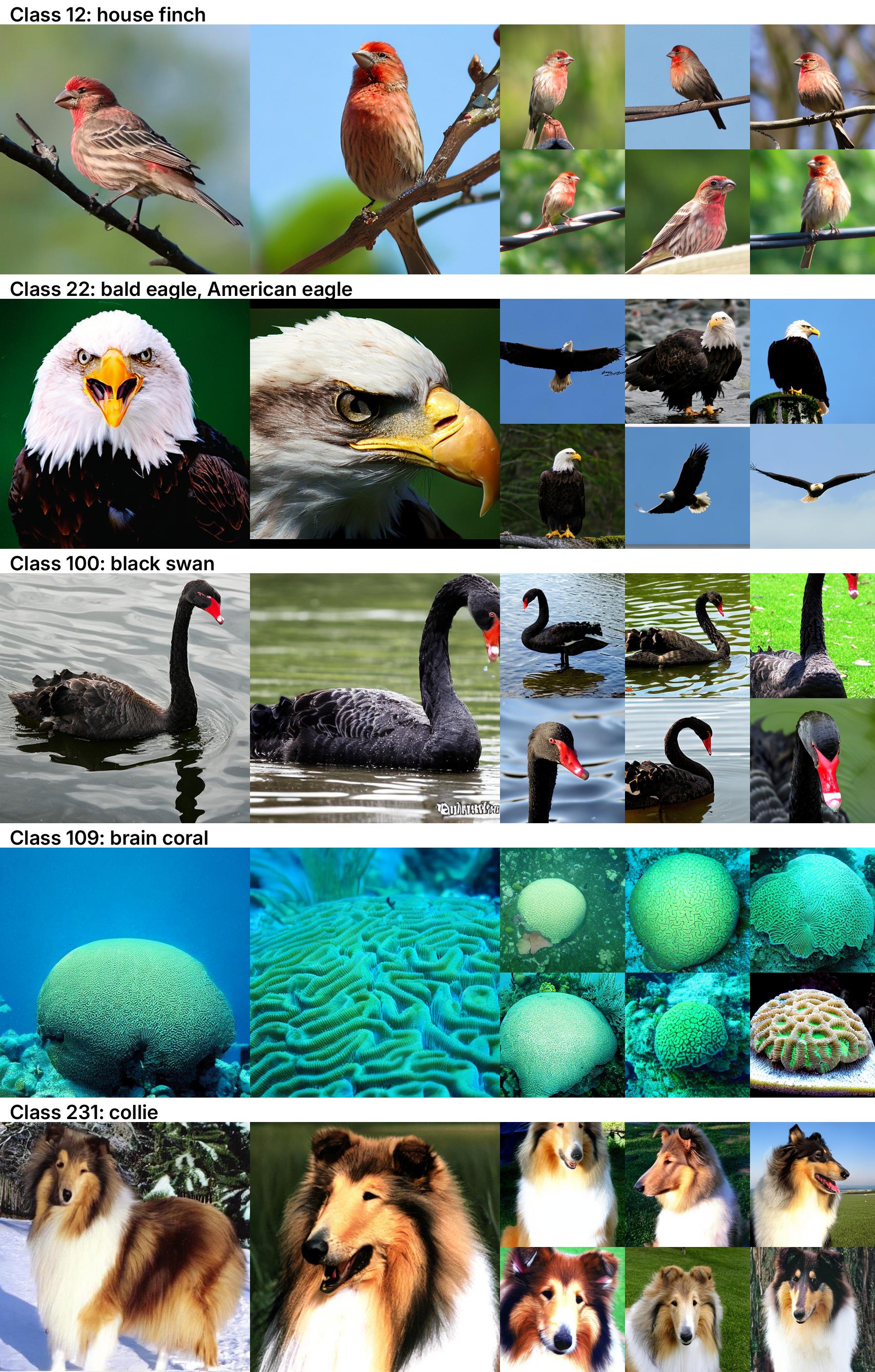} 
    \vspace{-0.1cm}
    \caption{Uncurated ImageNet samples at resolutions of $512 \times 512$ (left) and $256 \times 256$ (right), generated using UDT$^+$--XL/2 and UDT$^+$--XL/2 + REPA, respectively, with CFG ($\omega_{\mathrm{CFG}} = 4.0$).}
    \label{fig:XL_examples}    
\end{figure*}

\begin{figure*}[h]
    \centering
    \includegraphics[width=0.85\linewidth]{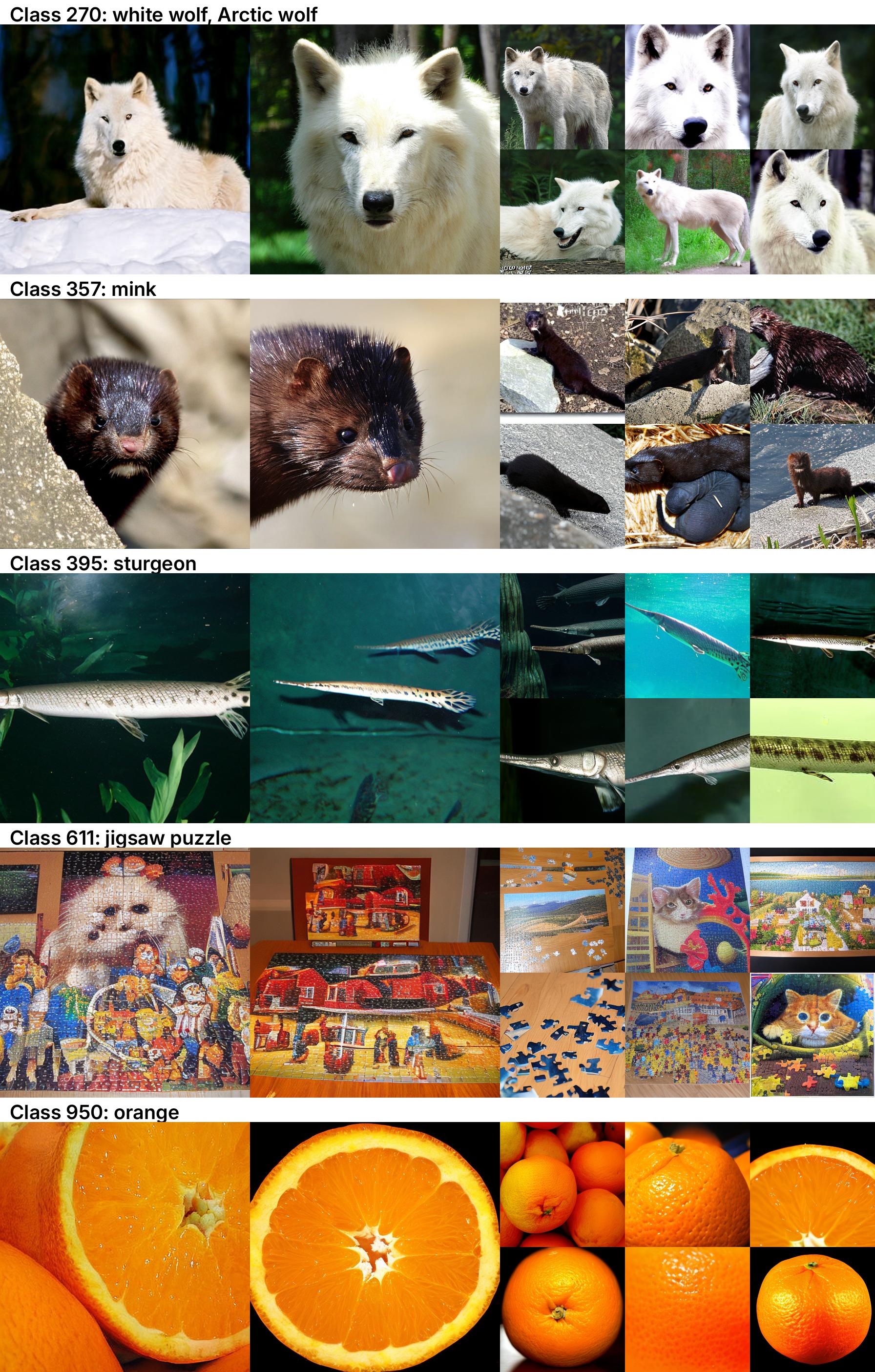} 
    \vspace{-0.1cm}
    \caption{Uncurated ImageNet samples at resolutions of $512 \times 512$ (left) and $256 \times 256$ (right), generated using UDT$^+$--XL/2 and UDT$^+$--XL/2 + REPA, respectively, with CFG ($\omega_{\mathrm{CFG}} = 4.0$).}
    \label{fig:XL_examples2}    
\end{figure*}

\end{document}